\documentclass{article}

\usepackage[nonatbib, final]{neurips_2022}
\usepackage[numbers]{natbib}
\usepackage[utf8]{inputenc} % allow utf-8 input
\usepackage[T1]{fontenc}    % use 8-bit T1 fonts

\usepackage{url}            % simple URL typesetting
\usepackage{booktabs}       % professional-quality tables
\usepackage{amsfonts}       % blackboard math symbols
\usepackage{nicefrac}       % compact symbols for 1/2, etc.
\usepackage{microtype}      % microtypography
\usepackage{xcolor}         % colors
\usepackage{amsmath}
\usepackage{algorithm}
\usepackage{algorithmic}
\usepackage{wrapfig}
\usepackage{graphicx}
\usepackage{caption}
\usepackage{subcaption}
\usepackage{wrapfig}
\usepackage{letltxmacro}

\usepackage{hyperref}
\hypersetup{colorlinks=true,linkcolor=black, citecolor=black}

\newcommand{\alg}{E3B}

\newcommand{\xxnote}[3]{}
\ifx\hidenotes\undefined
  \renewcommand{\xxnote}[3]{\color{#2}{#1: #3}}
\fi

\title{Exploration via Elliptical Episodic Bonuses}

\author{%
  Mikael Henaff \\
  Meta AI Research \\
  \texttt{mikaelhenaff@meta.com} \\
   \And
   Roberta Raileanu \\
   Meta AI Research \\
   \texttt{raileanu@meta.com}
   \AND
   \phantom{aaaaa}Minqi Jiang \\
   \phantom{aaaaa}University College London  \\
   \phantom{aaaaa}Meta AI Research\\
   \phantom{aaaaa}\texttt{meta@fb.com}   
   \And
   Tim Rockt{\"a}schel \\
   University College London \\
   \texttt{t.rocktaschel@cs.ucl.ac.uk}
}

\begin{document}

\maketitle

\begin{abstract}

In recent years, a number of reinforcement learning (RL) methods have been proposed to explore complex environments which differ across episodes.
In this work, we show that the effectiveness of these methods critically relies on a count-based episodic term in their exploration bonus.
As a result, despite their success in relatively simple, noise-free settings, these methods fall short in more realistic scenarios where the state space is vast and prone to noise. 
To address this limitation, we introduce \textbf{E}xploration via \textbf{E}lliptical \textbf{E}pisodic \textbf{B}onuses (\textbf{E3B}), a new method which extends count-based episodic bonuses to continuous state spaces and encourages an agent to explore states that are diverse under a learned embedding within each episode. The embedding is learned using an inverse dynamics model in order to capture controllable aspects of the environment.
Our method sets a new state-of-the-art across 16 challenging tasks from the MiniHack suite, without requiring task-specific inductive biases. \alg~also matches existing methods on sparse reward, pixel-based Vizdoom environments, and outperforms existing methods in reward-free exploration on Habitat, demonstrating that it can scale to high-dimensional pixel-based observations and realistic environments.

\end{abstract}

\section{Introduction}

%Exploration and generalization are two fundamental challenges in reinforcement learning. In recent years, methods which 

Exploration in environments with sparse rewards is a fundamental challenge in reinforcement learning (RL). In the tabular setting, provably optimal algorithms have existed since the early 2000s \cite{E3, Rmax}. 
%More recently, exploration has been studied in the context of deep RL, and a number of empirically successful methods have been proposed.
%These methods rely on intrinsically generated exploration bonuses that reward the agent for visiting states that are novel according to some measure, such as the likelihood of a state under a learned density model \cite{PseudoCounts}, the error of a forward dynamics model \cite{ICM} or the loss on a random prediction task \cite{RND}. These approaches have proven effective on hard exploration problems, as exemplified by the Atari games Montezuma's Revenge and PitFall~\cite{goexplore}. 
More recently, exploration has been studied in the context of deep RL, and a number of empirically successful methods have been proposed, such as pseudocounts \cite{PseudoCounts}, intrinsic curiosity modules (ICM) \cite{ICM}, and random network distillation (RND) \cite{RND}. These methods rely on intrinsically generated exploration bonuses that reward the agent for visiting states that are novel according to some measure. Different measures of novelty have been proposed, such as the likelihood of a state under a learned density model, the error of a forward dynamics model, or the loss on a random prediction task. These approaches have proven effective on hard exploration problems, as exemplified by the Atari games Montezuma's Revenge and PitFall~\cite{goexplore}. 

The approaches above are, however, designed for \textit{singleton} RL tasks, where the agent is spawned in the same environment in every episode. Recently, several studies have drawn attention to the fact that RL agents exhibit poor generalization across environments, and that even minor changes to the environment can lead to substantial degradation in performance \cite{overfitting_dl, illuminating_gen, dissection, pmlr-v97-cobbe19a, DBLP:journals/corr/abs-2111-09794}.
This has motivated the creation of benchmarks in the Contextual Markov Decision Process (CMDP) framework, where different episodes correspond to different environments that nevertheless share certain characteristics.
Examples of CMDPs include procedurally generated (PCG) environments \cite{gym_minigrid, minihack, NLE, obstacle_tower, procgen, deepmindlab, crafter, megaverse} or embodied AI tasks where the agent must generalize its behavior to unseen physical spaces at test time \cite{habitat19iccv, igibson, 3dworld, sapien}. 
%This has motivated the creation of benchmarks where different episodes correspond to different environments that nevertheless share certain characteristics: examples include procedurally generated (PCG) environments \cite{gym_minigrid, minihack, NLE, obstacle_tower, procgen, deepmindlab, crafter, megaverse} or embodied AI tasks where the agent must generalize its behavior to unseen physical spaces at test time \cite{habitat19iccv, igibson, 3dworld, sapien}. 
A number of methods have been proposed which have shown promising performance in PCG environments with sparse rewards, such as RIDE \cite{RIDE}, AGAC \cite{AGAC} and NovelD \cite{NovelD}. These methods propose different intrinsic reward functions, such as the change in representation in a latent space, the divergence between the predictions of a policy and an adversary, or the difference between random network prediction errors at two consecutive states. Although not presented as a central algorithmic feature, these methods also include a count-based bonus which is computed at the episode level. 

In this work, we take a closer look at exploration in CMDPs, where each episode corresponds to a different environment context.
We first show that, surprisingly, the count-based episodic bonus that is often included as a heuristic is in fact essential for good performance, and current methods fail if it is omitted.
%First, we clarify the different measures of novelty used in recent methods and show that they can be classified in two types: epistemic uncertainty, which measures novelty across all episodes and is analogous to the novelty measures used in singleton tasks, and what we call episodic uncertainty which measures novelty within a single episode. 
%We show that, surprisingly, the main driver of performance in recent methods is often not the stated algorithmic innovation, but rather the count-based episodic uncertainty term which is included as a heuristic. We then show that due to this dependence on a count-based term, existing methods fail on more complex tasks with irrelevant features or dynamic entities, where observations are rarely encountered more than once, unless task-specific inductive biases are applied. 
Furthermore, due to this dependence on a count-based term, existing methods fail on more complex tasks with irrelevant features or dynamic entities, where each observation is rarely seen more than once. We find that performance can be improved by counting certain features extracted from the observations, rather than the observations themselves. However, different features are useful for different tasks, making it difficult to design a feature extractor that performs well across all tasks.
% We investigate modifications to the count-based episodic bonus, and show that although performance can be improved by counting certain features extracted from the observations, rather than the observations themselves, different features are useful for different tasks, and it is challenging to design a feature extractor which performs well across all tasks. 

To address this fundamental limitation, we propose a new method, \alg, which uses an elliptical bonus \cite{Auer2002, Dani2008, LinUCB} at the episode level that can be seen as a natural generalization of a count-based episodic bonus to continuous state spaces, and that is paired with a self-supervised feature learning method using an inverse dynamics model. Our algorithm is simple to implement, scalable to large or infinite state spaces, and achieves state-of-the-art performance across 16 challenging tasks from the MiniHack suite \cite{minihack}, without the need for task-specific prior knowledge.
It also matches existing methods on hard exploration tasks from the VizDoom environment \cite{vizdoom}, and significantly outperforms existing methods in reward-free exploration on the Habitat embodied AI environment \cite{habitat19iccv, szot2021habitat}.
This demonstrates that E3B scales to rich, high dimensional pixel-based observations and real-world scenarios. Our code is available at \url{https://github.com/facebookresearch/e3b}. 

%To address this fundamental limitation, we propose a new method, \textbf{\alg}, which uses an \textbf{E}lliptical \textbf{E}pisodic \textbf{B}onus that can be seen as a natural generalization of a count-based bonus to continuous state spaces, and that is paired with a self-supervised feature learning method using an inverse dynamics model. Our algorithm is simple to implement, scalable to vast noisy observation spaces, and achieves state-of-the-art performance across 16 challenging tasks from the MiniHack suite, without the need for task-specific prior knowledge. It also matches existing methods on hard exploration tasks from the VizDoom environment, demonstrating that it scales well to high dimensional pixel-based observations. Our code is available at \url{web-link}. 

\section{Background}

\subsection{Contextual MDPs}

We consider a Contextual Markov Decision Process \footnote{\small{Technically, some of the environments we consider are Contextual Partially Observed MDPs (CPOMDPs), but we follow the convention in \cite{gen_review} and adopt the CMDP framework for simplicity. For CPOMDPs, we use recurrent networks or frame stacking to convert to CMDPs, as done in prior work \cite{DQN}}.} (CMDP~\cite{cmdp}) given by $\mathcal{M} = (\mathcal{S}, \mathcal{A}, \mathcal{C}, P, r, \mu_C, \mu_S)$ where $\mathcal{S}$ is the state space, $\mathcal{A}$ is the action space, $\mathcal{C}$ is the context space, $P$ is the transition function, $r$ is the reward function, $\mu_C$ is the distribution over contexts and $\mu_S$ is the conditional initial state distribution. At each episode, we sample a context $c \sim \mu_C$, an initial state $s_0 \sim \mu(\cdot | c)$, and subsequent states in the episode are sampled according to $s_{t+1} \sim P(\cdot | s_t, a_t, c)$. Let $d_\pi^c$ denote the distribution over states induced by following policy $\pi$ in context $c$. The goal is to learn a policy $\pi$ which maximizes the expected return over all contexts, i.e. $R = \mathbb{E}_{c \sim \mu_C, s \sim d_\pi^c, a \sim \pi(s)}[r(s, a)]$.
Examples of CMDPs include procedurally-generated environments, such as ProcGen~\cite{procgen}, MiniGrid \cite{gym_minigrid}, NetHack~\cite{NLE}, or MiniHack~\cite{minihack}, where each context $c$ corresponds to the random seed used to generate the environment; in this case, the number of contexts $|\mathcal{C}|$ is effectively infinite. Other examples include embodied AI environments \cite{habitat19iccv, szot2021habitat, 3dworld, igibson, sapien}, where the agent is placed in different simulated houses and must navigate to a location or find an object. In this setting, each context $c \in \mathcal{C}$ represents a house identifier and the number of houses $|\mathcal{C}|$ is typically between $20$ and $1000$. 
% More recently, CARL~\cite{carl} has been proposed as a benchmark for testing generalization in contextual MDPs. However, their focus is on using privileged information about the context $c$, which defines the physical properties of the current MDP, to improve generalization to new contexts. 
%More recently, CARL~\cite{carl} has been proposed as a benchmark for testing generalization in contextual MDPs. However, their focus is on using privileged information about the context $c$ to improve generalization, which we do not assume access to here. 
For an in-depth review of the literature on CMDPs and generalization in RL, see \cite{gen_review}. 

%In this work we will furthermore consider \textit{symbolic} CMDPs, a subclass of CMDPs where inputs are discrete symbols. Let $\mathcal{V}$ be a set of possible symbols. A symbolic MDPs state space has the form $\mathcal{S} = \mathcal{S}_1 \times \mathcal{S}_2 \times ... \times \mathcal{S}_M$, where each of the $\mathcal{S}_i$ are one-hot vectors of length $|\mathcal{V}|$. The popular MiniGrid and MiniHack suites fall in this category. 

\subsection{Exploration Bonuses}

If the environment rewards are sparse, learning a policy using simple $\epsilon$-greedy exploration may require intractably many samples. We therefore consider methods that augment the external reward function $r$ with an intrinsic reward bonus $b$. A number of intrinsic bonuses that encourage exploration in singleton (non-contextual) MDPs have been proposed, including pseudocounts \cite{PseudoCounts}, intrinsic curiosity modules (ICM) \cite{ICM} and random network distillation (RND) error \cite{RND}. At a high level, these methods define an intrinsic reward bonus that is high if the current state is different from the previous states visited by the agent, and low if it is similar (according to some measure).
% , where the similarity metric is typically based on learned observation embeddings.

%For example, Random Network Distillation (RND) \cite{RND} defines a bonus using the discrepancy between the prediction of a learned predictor network $f_\theta$ and a randomly initialized network $f_{\theta^{-}}$ at the current state:

%\begin{equation}
%    b(s_t) = \|f_\theta(s_t) - f_{\theta^{-}}(s_t) \|_2^2
%\end{equation}

% The intuition is that the error reflects the novelty of the current state $s_t$: it will be low if many states similar to $s_t$ have been encountered and high if not. The predictor function is trained online, using all the data which has been encountered by the agent. A number of other methods for exploration in singleton MDPs have been proposed, and are discussed in Section \ref{sec:related}.

%Another approach is to use the variance of an ensemble of dynamics models as a measure of novelty, the intuition being that the models will agree on states similar to the ones they have been trained on but not on unseen ones. The bonus then becomes:

%\begin{equation}
    %b(s, a) = \mathrm{Var}_{\theta_1, ..., \theta_E}[f_\theta(s, a)]
%\end{equation}

\begin{table}[h]
    \centering
    \begin{tabular}{ll}
        \hline
        Method \phantom{aaaakkk} & Exploration bonus \\
        \toprule
        \midrule
         \vspace{2mm}
         RIDE \cite{RIDE} & $\|\phi(s_{t+1}) - \phi(s_t)\|_2 \cdot
         \textcolor{blue}{1/\sqrt{N_e(s_{t+1})}}$ \\
         \vspace{2mm}
         NovelD \cite{NovelD} & $\Big[b_\mathrm{RND}(s_{t+1}) - \alpha \cdot b_\mathrm{RND}(s_t)\Big]_+ \cdot \textcolor{blue}{\mathbb{I}[N_e(s_{t+1}) = 1]}$ \\
         AGAC \cite{AGAC} & $D_\mathrm{KL}(\pi(\cdot | s_t)\| \pi_\mathrm{adv}(\cdot | s_t)) + \beta \cdot 
         \textcolor{blue}{1/\sqrt{N_e(s_{t+1})}}$
         \vspace{2mm} \\
         \hline
    \end{tabular}
    \vspace{1mm}
    \caption{Summary of recent exploration methods for procedurally-generated environments. Each exploration bonus has a count-based episodic term, marked in blue.}
    \vspace{-5mm}
    \label{tab:pcg_reward_bonuses}
\end{table}

More recently, several methods have been proposed for and evaluated on procedurally-generated MDPs. All use different exploration bonuses, which are summarized in Table \ref{tab:pcg_reward_bonuses}. RIDE \cite{RIDE} defines a bonus based on the distance between the embeddings of two consecutive observations, NovelD \cite{NovelD} uses a bonus based on the difference between two consecutive RND bonuses, and AGAC \cite{AGAC} uses the KL divergence between the predictions of the agent's policy and those of an adversarial policy trained to mimic it (see the original works for details). In addition, all three methods have a term in their bonus (marked in blue) which depends on $N_e(s_{t+1})$ -- the number of times $s_{t+1}$ has been encountered during the \textit{current episode}. Although presented as a heuristic, below we will show that without this count-based episodic term, all three methods fail to learn. This in turn limits their effectiveness in more complex, dynamic, and noisy environments.

\section{Importance and Limitations of Count-Based Episodic Bonuses}

%The original works introducing RIDE, AGAC and NovelD include the episodic terms which depend on $N_e(s_t)$ (the number of times the state $s_t$ has been encountered in the current episode) as heuristics. Here, we show that they are in fact essential for good performance. Figure \ref{fig:episodic_importance} shows NovelD, RIDE and AGAC on a MiniGrid task with and without the episodic bonus (experimental details can be found in Appendix \ref{appendix:experiments}). When the episodic bonus is removed, there is a substantial drop in performance for all three methods. 

%We now show that the count-based episodic bonuses in RIDE, AGAC and NovelD are in fact essential for good performance. 

\begin{figure}[h]
    \centering
    \includegraphics[width=\textwidth]{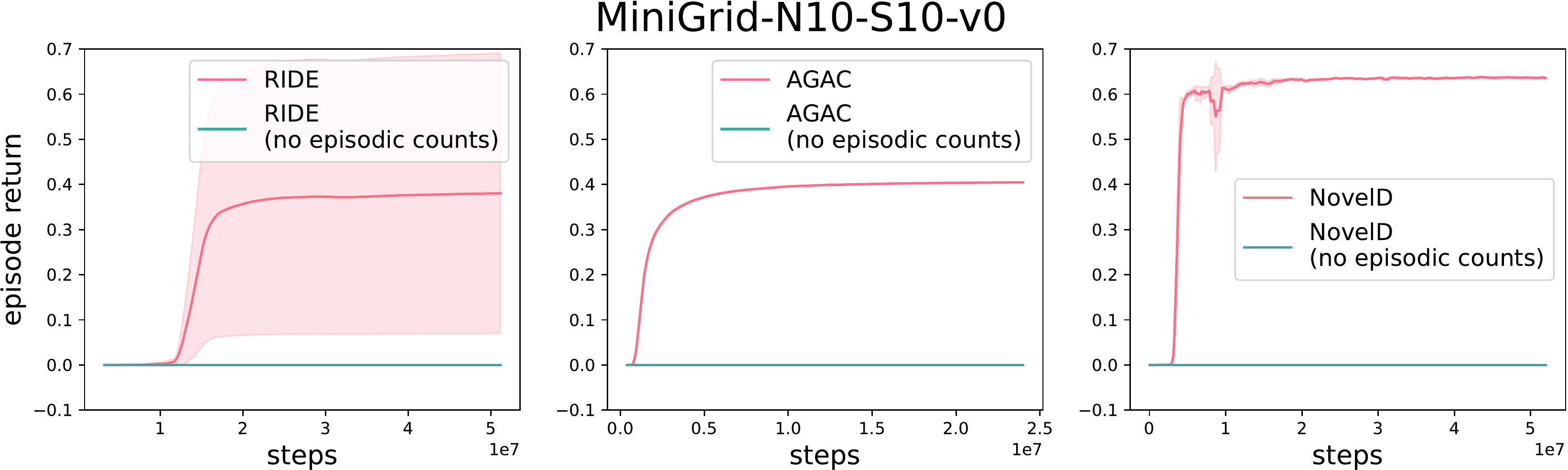} 
    \caption{Results for RIDE, AGAC and NovelD with and without the count-based episodic bonus, over $5$ random seeds (solid line indicates the mean, shaded region indicates one standard deviation). All three algorithms fail if the count-based episodic bonus is removed.}
    \label{fig:counts_ablation}
\end{figure}

We now discuss in more detail the importance and limitations of the count-based episodic bonuses used in RIDE, AGAC and NovelD, which depend on $N_e(s_t)$.
Figure \ref{fig:counts_ablation} shows results for the three methods with and without their respective count-based episodic terms, on one of the MiniGrid environments used in prior work. When the count-based terms are removed, all three methods fail to learn. Similar trends apply for other MiniGrid environments (see Appendix \ref{appendix:additional_minigrid_results}). This shows that the episodic bonus is in fact essential for good performance. 
%As we shall see in our experiments, these episodic bonuses are in fact essential for good performance. 

However, the count-based episodic bonus suffers from a fundamental limitation, which is similar to that faced by count-based approaches in general: if each state is unique, then $N_e(s_t)$ will always be $1$ and the episodic bonus is no longer meaningful. This is the case for many real-world applications. For example, a household robot's state as recorded by its camera might include moving trees outside the window, clocks showing the time or images on a television screen which are not relevant for its tasks, but nevertheless make each state unique.

%Many real-world applications have vast observation spaces, which renders most observations unique. For example, consider that we want to train a household robot to navigate to a certain room. The robot's observations may contain a lot of noise and irrelevant information such as humans moving around or clocks keeping track of time. Such environments would be challenging to explore if the agent gets the same intrinsic reward for each observation.
% There are many scenarios where observations can all be unique or close to it: for example, if a small amount of noise is added to the inputs, or if certain input dimensions contains irrelevent information or a time counter, or if there are many moving entities in the environment in addition to the agent, which leads to a combinatorially large number of possible configurations. 

Previous works \cite{RIDE, AGAC, NovelD} have used the MiniGrid test suite \cite{gym_minigrid} for evaluation, where observations are less noisy and do not typically contain irrelevant information. Thus, methods relying on episodic counts have been effective in these scenarios. 
However, in more complex environments such as MiniHack \cite{minihack} or with high-dimensional pixel-based observations, episodic count-based approaches can cease to be viable.

%A naive alternative could be to replace the episodic count-based term with another novelty bonus such as RND. However, obtaining an episodic bonus would require fitting a separate predictor network for each episode, which is impractical due to both the computational expense and the small number of samples.  
%Novelty bonuses based on neural density models or RND have been used instead of count-based bonuses in pixel-based setting for singleton MDPs. However, in that setting a single model is trained using all the data collected by the agent (i.e. aggregated across all episodes). Using such methods for an episodic bonus would require fitting a separate neural density model or predictor network for each episode, which is unfeasible due to both the computational expense and the small number of samples. 

%where they are fitted on all the data experienced by the agent, adopting this approach for episodic bonuses would require fitting a separate neural density model or predictor network for each episode, which is not practical due to both the computational expense and the small number of samples.

A possible alternative could be to design a function to extract relevant features from each state and feed them to the count-based episodic bonus. Specifically, instead of defining a bonus using $N_e(s_t)$, we could define the bonus using $N_e(\phi(s_t))$, where $\phi$ is a hand-designed feature extractor. For example, in the paper introducing the MiniHack suite \cite{minihack}, the RIDE implementation uses $\phi(s_t) = (x_t, y_t)$, where $(x_t, y_t)$ is the spatial location of the agent at time $t$. However, this approach relies heavily on task-specific knowledge.

\begin{wrapfigure}{r}{0.48\textwidth}
%    \vspace{-10mm}
    \vspace{-3mm}
   \begin{center}
     \includegraphics[width=0.48\textwidth]{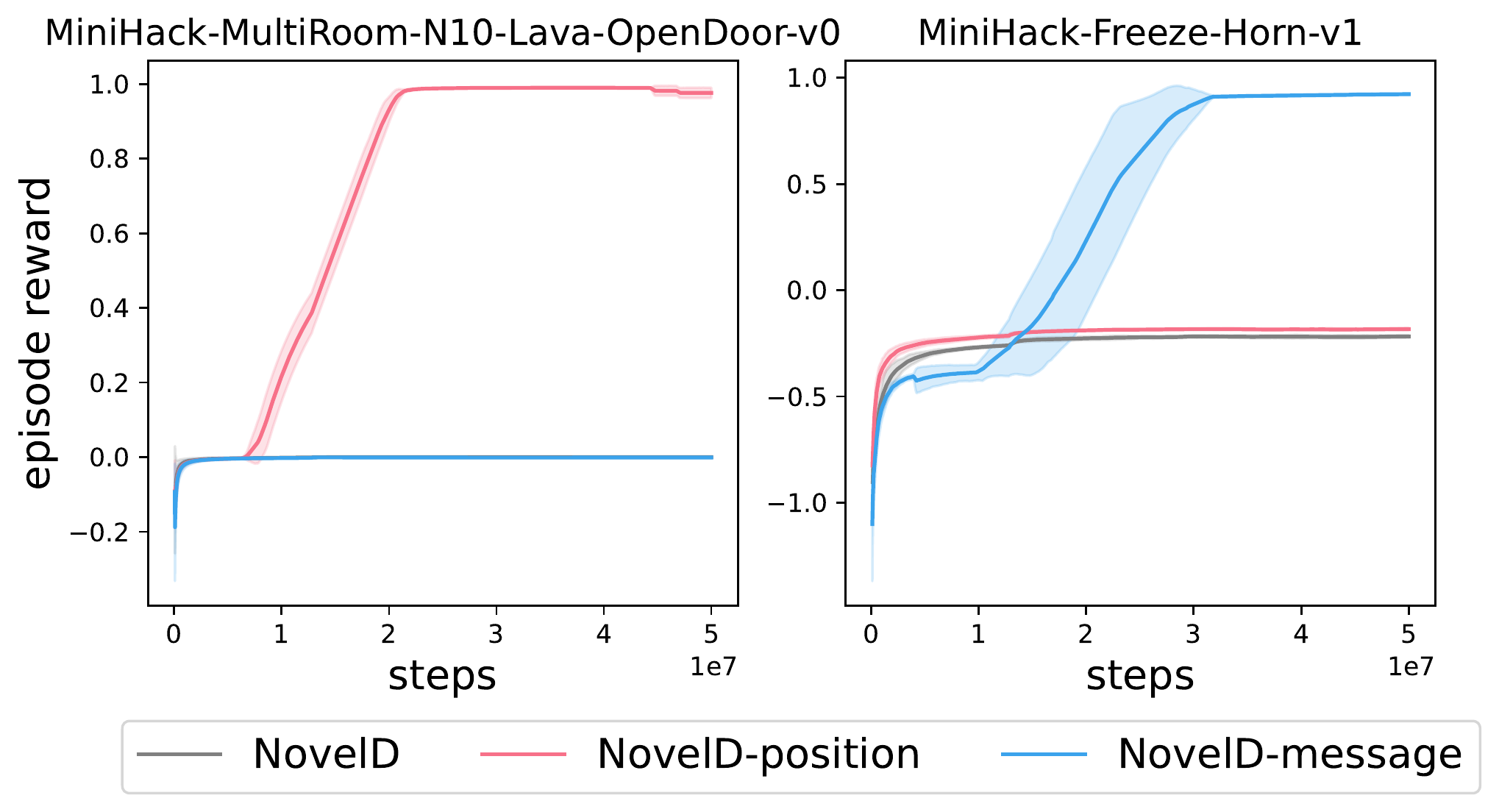}
   \end{center}
   \caption{Performance of NovelD with different feature extractors: entire observation, agent location, or environment message. Results are averaged over $5$ seeds and the shaded region represents one standard deviation.}
   \vspace{-3mm}
   \label{fig:small_table}
 \end{wrapfigure}

Figure \ref{fig:small_table} shows results on two tasks from the MiniHack suite for three NovelD variants (we decided to focus our study on variants of NovelD, since it was previously shown to outperform competing methods on MiniGrid \cite{NovelD}).
The first, \textsc{NovelD}, denotes the standard formulation which uses the bonus $\mathbb{I}[N_e(s_t)=1]$. The second, \textsc{NovelD-position}, uses the positional feature encoding described above, i.e. the episodic bonus is defined as $\mathbb{I}[N_e(\phi(s_t))=1]$ with $\phi(s_t) = (x_t, y_t)$. The third, \textsc{NovelD-message}, uses a feature encoding where $\phi(s_t)$ extracts the message portion of the state $s_t$ similarly to \cite{DBLP:journals/corr/abs-2202-08938} (both encodings are explained in more detail in Section \ref{sec:experiments}).
In contrast to the MiniGrid environments, here standard \textsc{NovelD} fails completely due to the presence of a time counter feature in the MiniHack observations, which makes each observation in the episode unique. Using the positional encoding enables \textsc{NovelD-position} to solve the \texttt{MultiRoom} task, but this method fails on the \texttt{Freeze} task. On the other hand, using the message encoding enables \textsc{NovelD-message} to succeed on the \texttt{Freeze} task, but it fails on the \texttt{MultiRoom} one. This illustrates that different feature extractors are effective on different tasks, and that designing one which is broadly effective is challenging.
Therefore, robust new methods which do not require task-specific engineering are needed.

\section{Elliptical Episodic Bonuses}

\begin{figure}
    \centering
    \includegraphics[width=\textwidth]{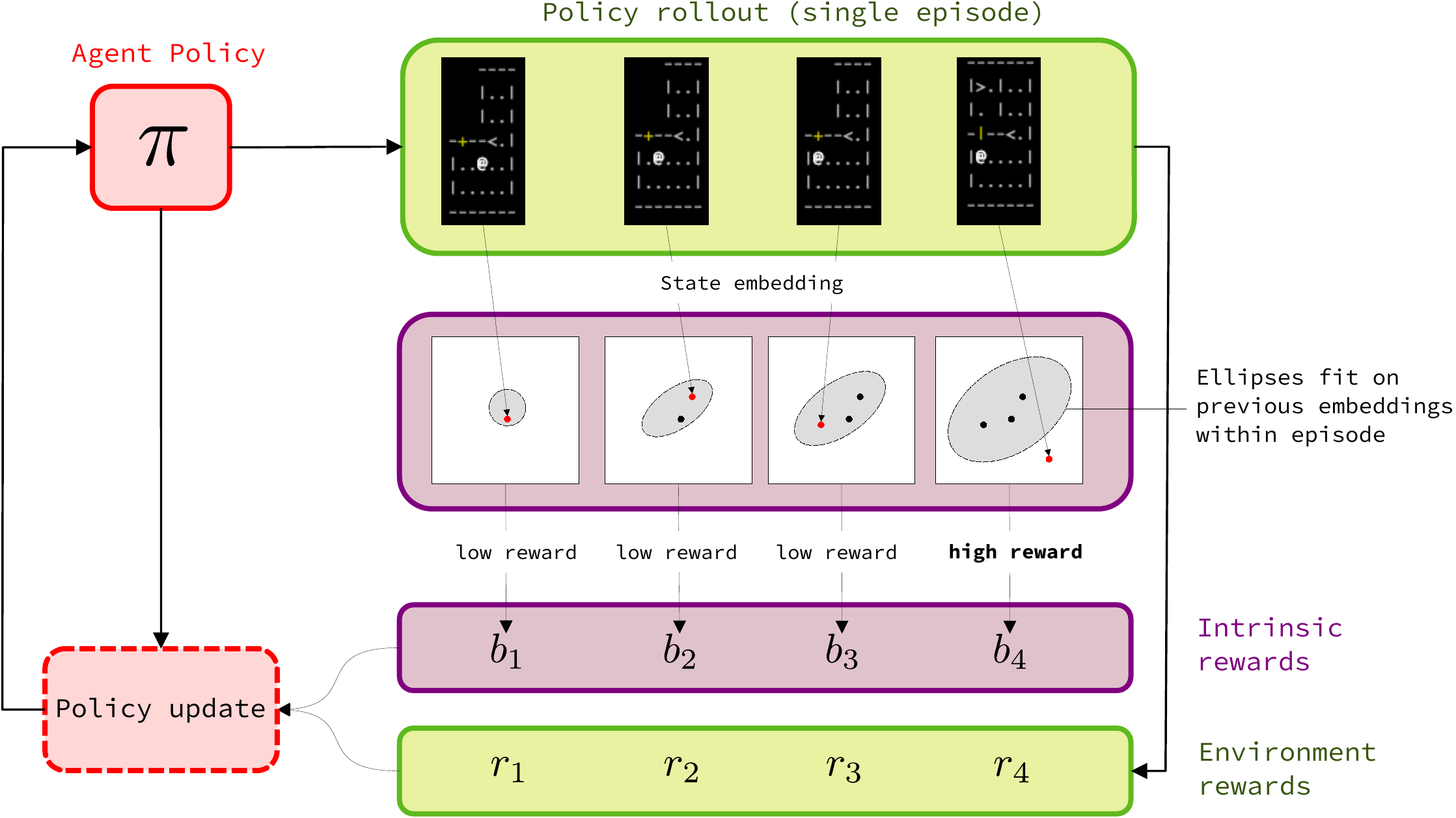}
    \caption{Overview of \alg. At each step in an episode, an ellipse is fit on the embeddings of previous states encountered within the episode. Intrinsic rewards are computed based on the position of the current state's embedding with respect to the ellipse: the further outside the ellipse, the higher the reward. These are combined with environment rewards, the policy is updated, and the process repeated. The state embeddings are learned using an inverse dynamics model.}
    \label{fig:summary}
\end{figure}

In this section we describe \textbf{Exploration via Elliptical Episodic Bonuses}, (\textbf{\alg}), our algorithm for exploration in contextual MDPs. It is designed to address the shortcomings of count-based episodic bonuses described above, with two aims in mind. First, we would like an episodic bonus that can be used with \textit{continuous} state representations, unlike the count-based bonus which requires discrete states. Second, we would like a representation learning method that only captures information about the environment that is relevant for the task at hand. The first requirement is met by using an elliptical bonus \cite{Auer2002, Dani2008, LinUCB}, which provides a continuous analog to the count-based bonus, while the second requirement is met by using a representation learned with an inverse dynamics model \cite{ICM, RIDE}. 

A summary of the method is shown in Figure \ref{fig:summary}. We define an intrinsic reward based on the position of the current state's embedding with respect to an ellipse fit on the embeddings of previous states encountered within the same episode. This bonus is then combined with the environment reward and used to update the agent's policy. The next two sections describe the elliptical bonus and embedding method in detail.
%The elliptical bonus and the embedding method are described in the next two sections. %and the full algorithm is detailed in Appendix \ref{appendix:alg_details}.%At each step, an ellipse is fit on the embeddings of previous observations seen within the current episode. An intrinsic reward bonus is computed based on the position of the current observation's embedding with respect to the ellipse. These intrinsic rewards are then combined with the environment rewards, and the policy is updated using policy gradient. Details of the elliptical bonus computation and the feature learning method are shown below.

%It relies on two key components: an elliptical bonus which measures episodic novelty, and a feature encoder learned using an inverse dynamics model which extracts controllable features of the environment from observations. 

\subsection{Elliptical Episodic Bonus}

Given a feature encoding $\phi$, at each time step $t$ in the episode the elliptical bonus $b$ is defined as follows:
\begin{align}
    b(s_t) = \phi(s_t)^\top C_{t-1}^{-1} \phi(s_t), \hspace{8mm}
    C_{t-1} = \sum_{i=1}^{t-1} \phi(s_i)\phi(s_i)^\top + \lambda I
\end{align}
\label{eq:elliptical_bonus}
Here $\lambda I$ is a regularization term to ensure that the matrix $C_{t-1}$ is non-singular, where $\lambda$ is a scalar coefficient and $I$ is the identity matrix. The reward optimized by the algorithm is then defined as $\bar{r}(s_t, a_t) = r(s_t, a_t) + \beta \cdot b(s_t)$, where $r(s_t, a_t)$ is the extrinsic reward provided by the environment and $\beta$ is a scalar term balancing the tradeoff between exploration and exploitation. 

\textbf{Intuition.} One perspective is that the elliptical bonus is a natural generalization of a count-based episodic bonus \cite{PCPG}. To see this, observe that if the problem is tabular and $\phi$ is a one-hot encoding of the state, then $C_{t-1}$ will be a diagonal matrix whose entries contain the counts corresponding to each state encountered in the episode. Its inverse $C_{t-1}^{-1}$ will also be a diagonal matrix whose entries are inverse state visitation counts, and the bilinear form $\phi(s_t)^\top C_{t-1}^{-1} \phi(s_t)$ reads off the entry corresponding to the current state $s_t$, yielding a bonus of $1/N_e(s_t)$.

For a more general geometric interpretation, if $\phi(s_0),...,\phi(s_{t-1})$ are roughly centered at zero, then $C_{t-1}$ can be viewed as their unnormalized covariance matrix. Now consider the eigendecomposition $C_{t-1} = U^\top \Lambda U$, where $\Lambda$ is the diagonal matrix whose entries are the eigenvalues $\lambda_1, ..., \lambda_n$ (these are real since $C_{t-1}$ is symmetric). Letting $z = U\phi(s_t) = (z_1, ..., z_n)$ be the set of coordinates of $\phi(s_t)$ in the eigenspace of $C_{t-1}$, we can rewrite the elliptical bonus as:
\begin{align*}
    b(s_t) = z^\top \Lambda^{-1} z = \sum_{i=1}^n \frac{z_i^2}{\lambda_i}
\end{align*}
The bonus increases the more $\phi(s_t)$ is aligned with the eigenvectors corresponding to smaller eigenvalues of $C_{t-1}$ (directions of low data density), and decreases the more it is aligned with eigenvectors corresponding to larger eigenvalues (directions of high data density). An illustration for $n=2$ is shown in Figure \ref{fig:elliptical2d} of Appendix \ref{appendix:alg_details}. In our experiments, $n$ is typically between $256$ and $1024$.

%\mikael{TODO: give better intuitions here, maybe add some figures illustrating ellipses}
%There are several perspectives which provide intuition for the elliptical bonus. One perspective is that the matrix $C_t$ defines an ellipsoid centered at the origin, whose axes are proportional to the eigenvalues of $C_t$ and whose direction is determined by its eigenvectors. The bonus is then the width of the ellipsoid in the direction of the point $\phi(s_t)$. 

%A second perspective on the elliptical bonus is that it is a natural generalization of a count-based bonus. To see this, observe that if the problem is tabular and $\phi$ is a one-hot encoding of the state, then $C_{t-1}$ will be a diagonal matrix whose entries contain the counts corresponding to each state encountered in the episode. Its inverse $C_{t-1}^{-1}$ will also be a diagonal matrix whose entries are inverse counts, and the bilinear form $\phi(s_t)^\top C_{t-1}^{-1} \phi(s_t)$ reads off the entry corresponding to the current state $s_t$, yielding a bonus of $1/N(s_t)$.

%The elliptical bonus is related to the Mahalanobis distance \cite{mahalanobis1936generalized} which uses a similar bilinear form. However, the Mahalanobis distance would normalize the matrix $C_{t-1}$ in equation \ref{eq:elliptical_bonus} by the number of observations $t-1$ in the episode, whereas the elliptical bonus does not.The elliptical bonus thus tends to decrease with the number of observations, similarly to the count-based bonus. 

\textbf{Efficient Computation.} The matrix $C_{t-1}$ needs to be inverted at each step $t$ in the episode, an operation which is cubic in the dimension of $\phi$ and may thus be expensive. To address this, we use the Sherman-Morrison matrix identity \cite{Sherman-Morrison-1950} to perform fast rank-$1$ updates in quadratic time:

\begin{equation*}
C_t^{-1} = 
    \begin{cases}
    \frac{1}{\lambda}I \phantom{------------} \text{ if } t=0 \\
    C_{t-1}^{-1} - \frac{C_{t-1}^{-1}\phi(s_{t})\phi(s_{t})^\top C_{t-1}^{-1\top}}{1 + \phi(s_{t})^\top C_{t-1}^{-1}\phi(s_{t})} \phantom{-}\text{ if } t \geq 1
    \end{cases}
\end{equation*} 

This results in an approximately $3\times$ speedup over na\"{\i}ve matrix inversion (details in Appendix \ref{appendix:rank1_exp}).

%A comparison of \alg's speed with and without the rank-$1$ update is given in Appendix \ref{appendix:rank1_exp}.

%\begin{align*}
%C_t^{-1} &= \frac{1}{\lambda}I    \\
%C_{t+1}^{-1} &= C_t^{-1} - \frac{C_t^{-1}\phi(s_{t+1})\phi(s_{t+1})^\top C_t^{-1}^\top}{1 + \phi(s_{t+1})^\top C_t^{-1}\phi(s_{t+1})} \hspace{10mm} \mbox{ for } t \geq 1
%\end{align*}

\subsection{Learned Feature Encoder}

Any feature learning method could in principle be used to learn $\phi$. Here we use the inverse dynamics model approach proposed in \cite{ICM}, which trains a model $g$ along with $\phi$ to map each pair of consecutive embeddings $\phi(s_t), \phi(s_{t+1})$ to a distribution over actions $a_t$ linking them. In our setup, $\phi$ is separate from the policy network. The $g$ model is trained jointly with $\phi$ using the following per-sample loss:

\begin{equation*}
    \ell(s_t, a_t, s_{t+1}; \phi, g) = -\log p(a_t | g(\phi(s_t), \phi(s_{t+1})))
\end{equation*}

The motivation is that the mapping $\phi$ will discard information about the environment which is not useful for predicting the agent's actions. Previous work \cite{ICM} has shown that this can make learning more robust to random noise or other parts of the state which are not controllable by the agent. 
%In our experiments, we compare this to other approaches such as using the policy network \cite{ACB} or random networks \cite{pathak18largescale, PCPG} to produce state embeddings. 

\subsection{Full Algorithm}

Putting all of these together, the full algorithm is given below.

\newlength{\commentindent}
\setlength{\commentindent}{.5\textwidth}
\makeatletter
\renewcommand{\algorithmiccomment}[1]{\unskip\hfill\makebox[\commentindent][l]{//~#1}\par}
\LetLtxMacro{\oldalgorithmic}{\algorithmic}
\renewcommand{\algorithmic}[1][0]{%
  \oldalgorithmic[#1]%
  \renewcommand{\ALC@com}[1]{%
    \ifnum\pdfstrcmp{##1}{default}=0\else\algorithmiccomment{##1}\fi}%
}
\makeatother

\begin{algorithm}[h!]
\caption{Exploration via Episodic Elliptical Bonuses (\alg)}
\begin{algorithmic} 
%\REQUIRE $n \geq 0 \vee x \neq 0$
\STATE Initialize policy $\pi$, feature encoder $\phi$ and inverse dynamics model $f$.
\WHILE{not converged}
\STATE Sample context $c \sim \mu_C$ and initial state $s_0 \sim \mu_S(\cdot | c)$
\STATE Initialize inverse covariance matrix: $C_0^{-1} = \frac{1}{\lambda} I$
\FOR{$t = 0, ..., T$}
\STATE $a_t \sim \pi(\cdot | s_t)$  \COMMENT{Sample action}
\STATE $s_{t+1}, r_{t+1} \sim P(\cdot | s_t, a_t)$  \COMMENT{Step through environment}
\STATE $b_{t+1} = \phi(s_{t+1})^\top C_t^{-1} \phi(s_{t+1})$ \COMMENT{Compute bonus}
\STATE $u = C_t^{-1}\phi(s_{t+1})$
\STATE $C_{t+1}^{-1} = C_t^{-1} - \frac{1}{1 + b_{t+1}} uu^\top$ \COMMENT{Update inverse covariance matrix}
\STATE $\bar{r}_{t+1} = r_{t+1} + \beta b_{t+1}$
\ENDFOR
\STATE Perform policy gradient update on $\pi$ using rewards $\bar{r}_1, ..., \bar{r}_T$.
\STATE Update $\phi$ and $g$ using $\{(s_t, a_t, s_{t+1})\}_{t=0}^{T-1}$ to minimize the loss: 
\begin{equation*}
\ell = -\log(p(a_t | f(\phi(s_t), \phi(s_{t+1}))))
\end{equation*}
%$\ell = -\log(p_f(a_t | \phi(s_t), \phi(s_{t+1})))$
\ENDWHILE
\end{algorithmic}
\end{algorithm}

\section{Experiments}
\label{sec:experiments}

\subsection{MiniHack Suite}

In order to probe the capabilities of existing methods and evaluate \alg, we seek CMDP environments which exhibit challenges associated with realistic scenarios, such as sparse rewards, noisy or irrelevant features, and large state spaces. For our first experimental testbed, we opted for the procedurally generated tasks from the MiniHack suite \cite{minihack}, which is itself based on the NetHack Learning Environment \cite{NLE}.
%As our first evaluation testbed, we use procedurally generated tasks from the MiniHack suite \cite{minihack}, which is itself based on the NetHack Learning Environment \cite{NLE}. 
NetHack is a notoriously challenging roguelike video game where the agent must navigate through procedurally generated dungeons to recover a magical amulet. 
The MiniHack tasks contain numerous challenges such as finding and using magical objects, navigating through levels while avoiding lava, and fighting monsters. Furthermore, rewards are sparse and as detailed below, the state representation contains a large amount of information, only some of which is relevant for a given task. 

%The agent is faced with numerous challenges, such as finding and using magical objects, fighting monsters, navigating through large levels and avoiding traps and lava. 
%The MiniHack tasks contain irrelevant features, dynamic entities (such as monsters), and only provide a reward upon completion of a complex task, which makes them a good testbed for our purposes.
%The MiniHack tasks are designed to test precise agent capabilities, such as its ability to solve these different challenges in isolation or in combination. %From an RL standpoint, MiniHack and NetHack pose challenges associated with exploration, generalization, stochastic dynamics, partial observability, and large action spaces.

For all our experiments we use the Torchbeast \cite{torchbeast} implementation of IMPALA \cite{IMPALA} as our base RL algorithm. For certain skill-based tasks, we restricted the action space to the necessary actions for solving the task at hand, since we found that none of the methods were able to make progress with the full action space (see Appendix \ref{appendix:action_restriction}).
See Appendix \ref{appendix:env_details} for environment details and \ref{appendix:minihack} for other experiment details.
%Full details on the experiments and environments are in Appendix \ref{appendix:experiments} and \ref{appendix:env_details}.%, and details on the tasks in Appendix \ref{appendix:env_details}.

\subsubsection{Modalities for Episodic Bonus}

\begin{wrapfigure}{r}{0.42\textwidth}
\centering
    \centerline{\includegraphics[width=0.42\textwidth]{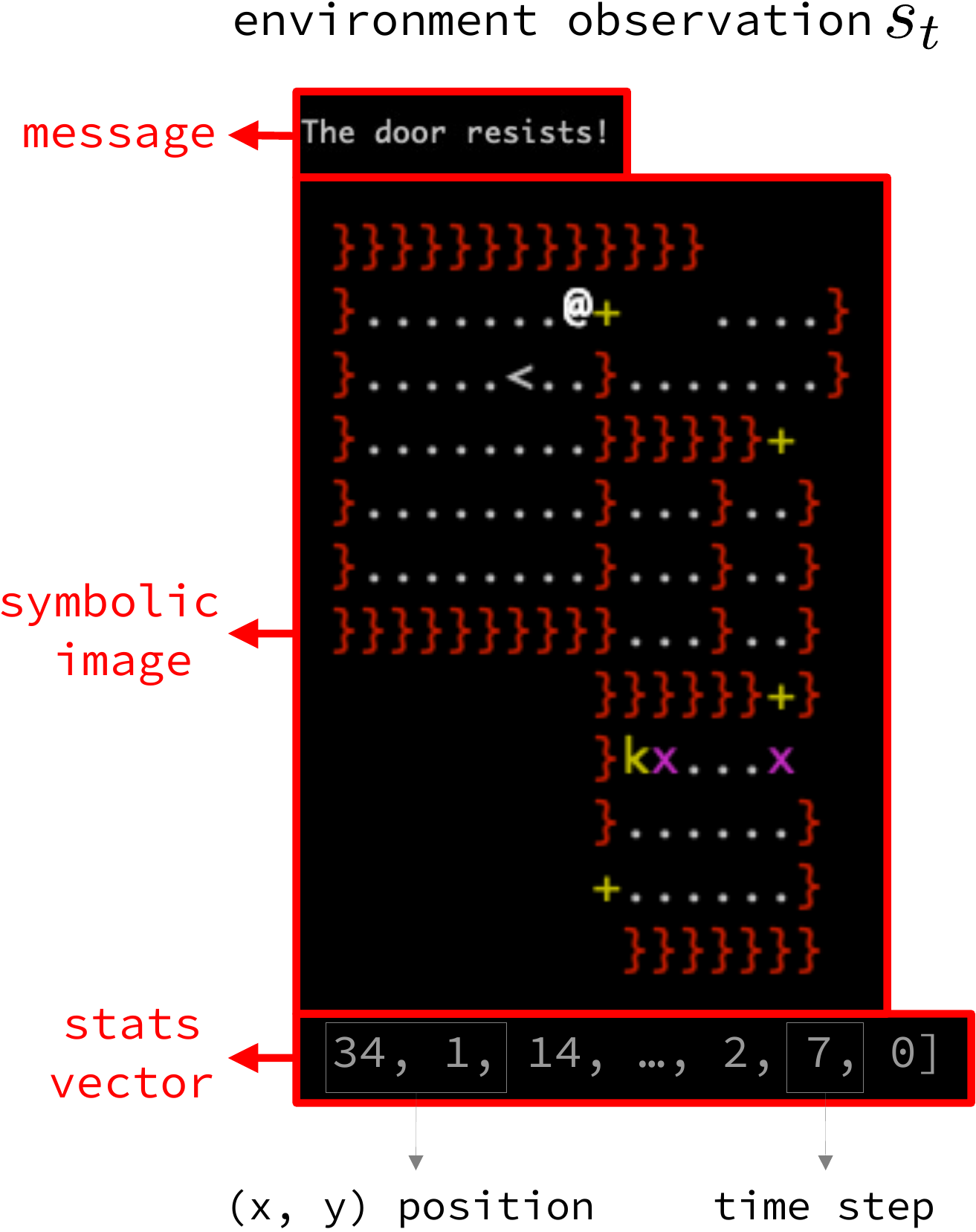}}
  \caption{Observation for MiniHack }
  \vspace{-5mm}
  \label{fig:modalities}
\end{wrapfigure}

The MiniHack environments provide an observation at each step that includes three different modalities: i) a symbolic image, which indicates the location of the agent, monsters or other entities, objects and different types of terrain (such as walls, lava, water); ii) a statistics vector which indicates the agent's current $(x, y)$ location, hit points, time step $t$, and other features such as strength, dexterity and constitution, and iii) a textual message which gives different types of feedback about the environment ("you see a key of master thievery", "the wall is solid rock"). These are illustrated in Figure \ref{fig:modalities}.

We now draw attention to some important subtleties regarding how existing methods implement count-based episodic bonuses---as we will see, these can have a large impact on performance. The works of \cite{RIDE, AGAC, NovelD}, which use the MiniGrid suite \cite{gym_minigrid} for evaluation, use a hash table whose keys are the full observations (in this case symbolic images) encountered by the agent. The occurrences of these observations are then counted to compute the episodic bonus. The baselines run in the MiniHack tasks \cite{minihack}, on the other hand, use a hash table whose keys are the $(x, y)$ positions of the agent, which are extracted from the observation and are then similarly counted to compute the episodic bonus. 

%Which of these is preferable? On one hand, using the full observation does not discard any potentially useful information, but if there are irrelevant features in the observations, similar states might be counted as separate, which can make the count-based episodic bonus meaningless. On the other hand, using the $(x,y)$ positions might be useful for navigation-based tasks, and can discard distracting information, but might discard useful information for tasks which are not navigation based.
%\vspace{-10mm}
In order to understand the effect of these different input modalities on count-based episodic bonuses, we consider three variants of the NovelD algorithm, in addition to the original formulation. These methods all have a count-based episodic bonus of the form $\mathbb{I}[N_e(\phi(s_t))=1]$, with different choices of $\phi$ detailed below.

\underline{\textsc{NovelD}}: in the original version of the algorithm, $\phi$ is the identity. Here the input to the count-based episodic bonus is the full state, namely the concatenation of the symbol image, the message and the stats vector. This makes no assumptions about which part of the state is most useful. 

\underline{\textsc{NovelD-position}}: in this variant, $\phi$ extracts the $(x, y)$ position of the agent from the statistics vector, which reflects a strong inductive bias that the task is navigation-based.
    
\underline{\textsc{NovelD-image}}: in this variant, $\phi$ extracts the symbol image only. This reflects an inductive bias that the message and the stats vector are not useful since they are discarded.
    
\underline{\textsc{NovelD-message}}: in this variant, $\phi$ extracts the message only. This reflects an inductive bias that the messages are important but the stats and symbolic images are not.  

\begin{figure}
    \centering
    \includegraphics[width=\textwidth]{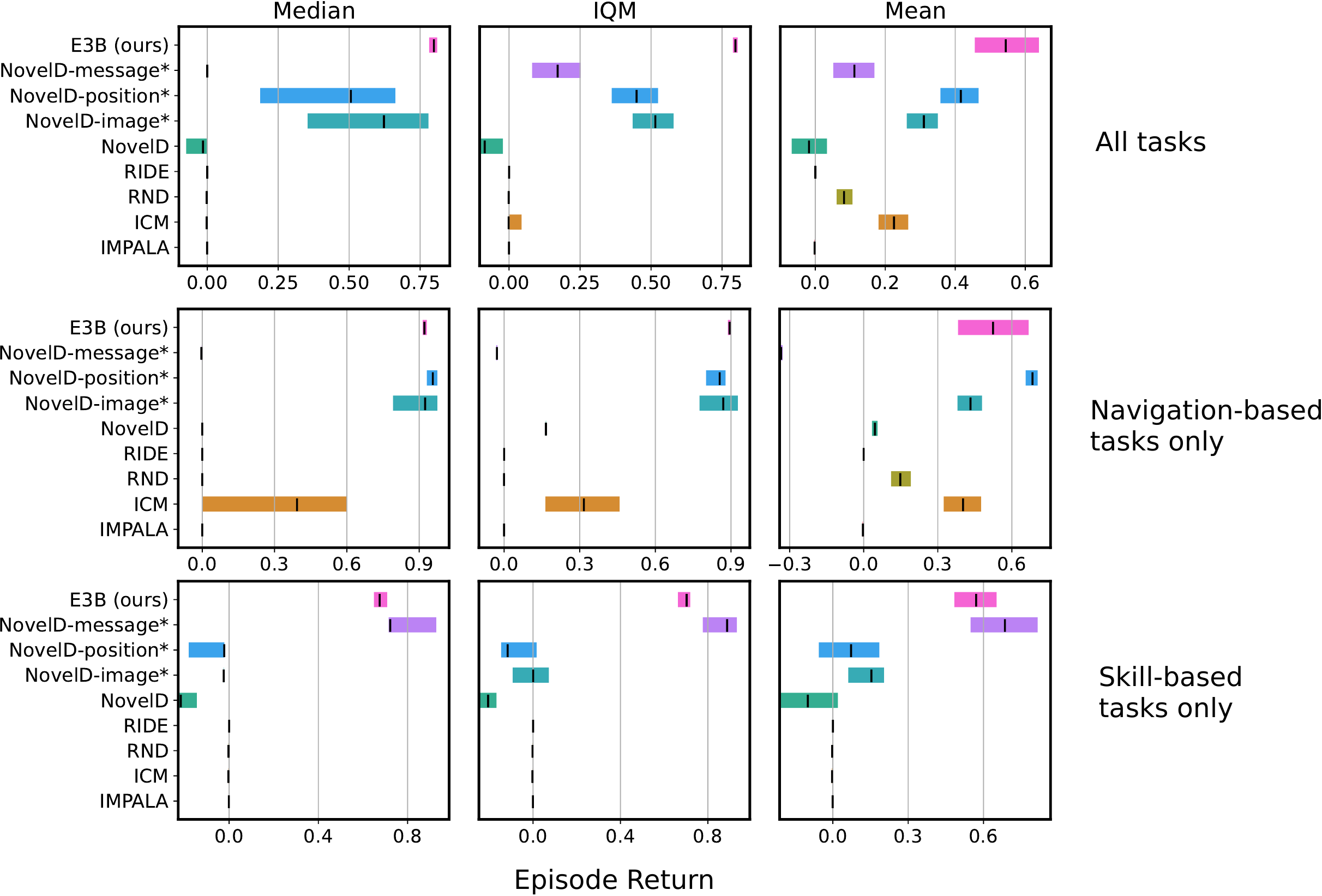}
    \caption{Aggregate results over 16 tasks from the MiniHack environment. Bars represent $95\%$ confidence intervals computed using stratified bootstrapping with $5$ random seeds. Methods marked with $^*$ use task-specific prior knowledge.}
    \label{fig:aggregate}
\end{figure}

% \begin{itemize}
%     \item \textbf{NovelD}: here the count-based episodic bonus is computed by counting the number of times the full observation has been encountered by the agent. The full observation is the concatenation of the symbol image, the message and the stats vector. This makes no assumptions about which part of the observation is most useful. 

%     \item \textbf{NovelD-position}: here the count-based episodic bonus is computed by counting the number of times each $(x, y)$ position has been encountered. Note that this requires extracting the position from the observation, which may require feature engineering, and reflects a strong inductive bias that the task is navigation-based.
    
%     \item \textbf{NovelD-image}: here the count-based episodic bonus is computed by counting the number of times each symbol image has been encountered by the agent. This reflects an inductive bias that the message and the stats vector are not useful since it discards them.
    
%     \item \textbf{NovelD-message}: here the count-based episodic bonus is computed by counting the number of times each message has been encountered by the agent. This reflects an inductive bias that the messages are important but the stats and symbolic images are not.  
% \end{itemize}

For \alg, we feed the full state to the algorithm and do not make any assumptions about which part is useful for the task at hand. Despite the lack of task-specific inductive biases, our method is still able to extract the relevant features for each task, thus outperforming the other exploration approaches (or matching their performance when prior knowledge of the task is used). 

\subsubsection{Results on MiniHack}

Aggregate results for IMPALA, RND, RIDE, ICM, NovelD (with the three variants described above) and \alg~over $16$ sparse reward tasks from the MiniHack suite are shown in Figure \ref{fig:aggregate}. 
Out of $16$ environments, $8$ are based on the MiniGrid suite, but use the MiniHack interface and observation space (environment details are included in Appendix \ref{appendix:env_details}). We used the performance metrics and bootstrapping protocol from \cite{rliable} to compute confidence intervals, which are more informative than simple point estimates. We see that over all tasks (top row), \alg~outperforms all other methods by a significant margin across all three performance metrics. 

Standard \textsc{NovelD} performs poorly due to the fact that the MiniHack observations contain a time counter in the statistics vector (see Figure \ref{fig:modalities}), which makes each observation in the episode unique and hence renders the count-based episodic bonus meaningless. The three NovelD variants which use different features extracted from the state for the count-based bonus perform better. Figure \ref{fig:aggregate} shows aggregate performance across a subset of $9$ navigation-based tasks (middle row) and $7$ skill-based tasks (bottom row) (see Appendix \ref{appendix:env_details} for the task breakdown). On the navigation-based tasks, \textsc{NovelD-position} has excellent performance, since visiting a large number of different $(x,y)$ locations is closely aligned with the true task reward. However, \textsc{NovelD-position} fails on all the skill-based tasks, resulting in poor performance overall. We observe an opposite trend for the \textsc{NovelD-message} variant, which performs very well on the skill-based tasks, but poorly on the navigation-based ones. 

This highlights that although certain inductive biases can help for certain tasks, it is difficult to find one which performs well across all of them. Our method, \alg, performs well on both the navigation-based tasks and the skill-based tasks, resulting in superior performance overall. It is worth noting that \alg ~does not require any task-specific engineering: the exact same algorithm is run for all tasks, and it uses the unprocessed states. Results for individual tasks, along with more analysis and discussion, can be found in Appendix \ref{appendix:additional_minihack_discussion}.

\subsubsection{Ablation Experiments}
 \begin{wrapfigure}{r}{0.5\textwidth}
    \vspace{-22mm}
   \begin{center}
     \includegraphics[width=0.48\textwidth]{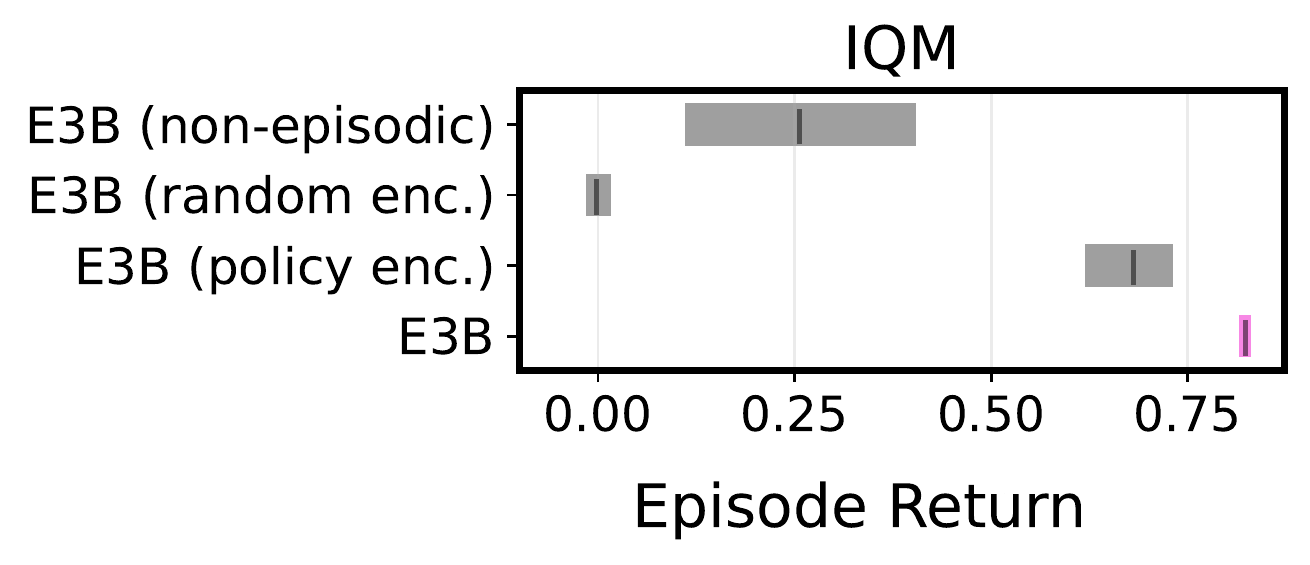}
   \end{center}
   \caption{MiniHack ablations, bars represent $95\%$ confidence intervals using stratified bootstrapping.}   
%   \caption{Interquartile Mean (IQM) performance over $16$ MiniHack tasks. Bars represent $95\%$ confidence intervals based on stratified bootstrapping.}
   \vspace{-4mm}
   \label{fig:ablations}
 \end{wrapfigure}
We next report results for ablation experiments measuring the effects of different algorithmic components of \alg, namely the feature encoding $\phi$ and the episodic nature of the bonus. 
Results are shown in Figure \ref{fig:ablations}. 
\alg~(random enc.) indicates \alg~where $\phi$ is a randomly initialized network which is kept fixed throughout training. 
 \alg~(policy enc.) indicates \alg~where the weights of $\phi$ are tied to those of the policy network (with the last layer producing action probabilities removed). \alg~(non-episodic) indicates \alg~where the elliptical bonus is computed using observations across \textit{all} timesteps in the lifetime of the agent, and not just the current episode. 
 %This variant is conceptually similar to other exploration methods for singleton MDPs in that is measures lifelong novelty rather than novelty within the current episode alone. 
 All three variants perform significantly worse than \alg, which uses an inverse dynamics model encoding for $\phi$ and an episodic bonus. This highlights that both the inverse dynamics model and episodic bonus are important for success.

\subsection{Pixel-Based VizDoom}

As our second evaluation testbed, we used the sparse reward, pixel-based VizDoom \cite{vizdoom} environments used in prior work \cite{ICM, RIDE}. Although these are singleton MDPs, they still constitute challenging exploration problems and probe whether our method scales to continuous high-dimensional pixel-based observations. Results comparing \alg~to RIDE, ICM and IMPALA on three versions of the task are shown in Figure \ref{fig:vizdoom} (hyperparameters can be found in Appendix \ref{appendix:vizdoom_hps}). IMPALA succeeds on the dense reward task but fails on the two sparse reward ones. \alg~is able to solve both versions of the sparse reward task, similar to RIDE and ICM.

We emphasize that these are \textit{singleton} MDPs, where the environment does not change from one episode to the next. Therefore, it is unsurprising that ICM, which was designed for singleton MDPs, succeeds in this task. RIDE is also able to solve the task, consistent with results from prior work \cite{RIDE}. The fact that \alg~also succeeds provides evidence of its robustness and its applicability to settings with pixel-based observations. 

\begin{figure}[h]
    \centering
    \includegraphics[width=\textwidth]{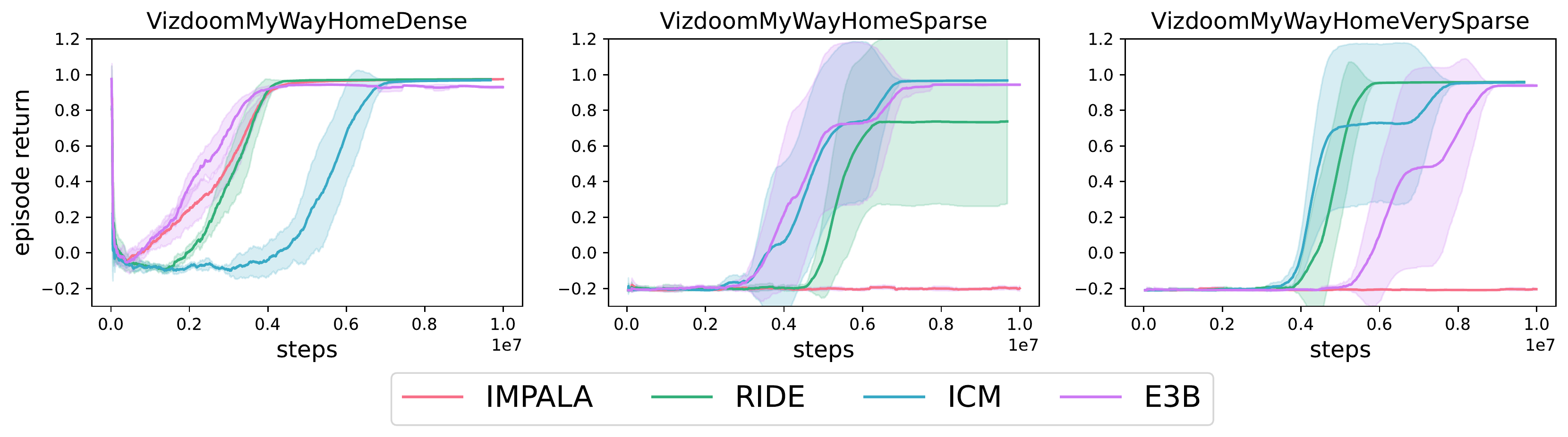}
    \caption{Results on pixel-based Vizdoom tasks with dense and sparse rewards. Results are averaged over $5$ random seeds, shaded region indicates one standard deviation.}
    \label{fig:vizdoom}
\end{figure}

%To test how our elliptical bonus performs in high-dimensional pixel-based setting, we additionally performed experiments on the sparse reward VizDoom environments used in \cite{ICM, RIDE}. 

%\mikael{I remember we discussed some extra parts for this section but I forgot what all they were, could you remind me?}

\subsection{Reward-free Exploration on Habitat}

As our third experimental setting, we investigate reward-free exploration in Habitat \cite{habitat19iccv, szot2021habitat}. 
 \begin{wrapfigure}{r}{0.36\textwidth}
%    \vspace{-10mm}
   \begin{center}
     \includegraphics[width=0.36\textwidth]{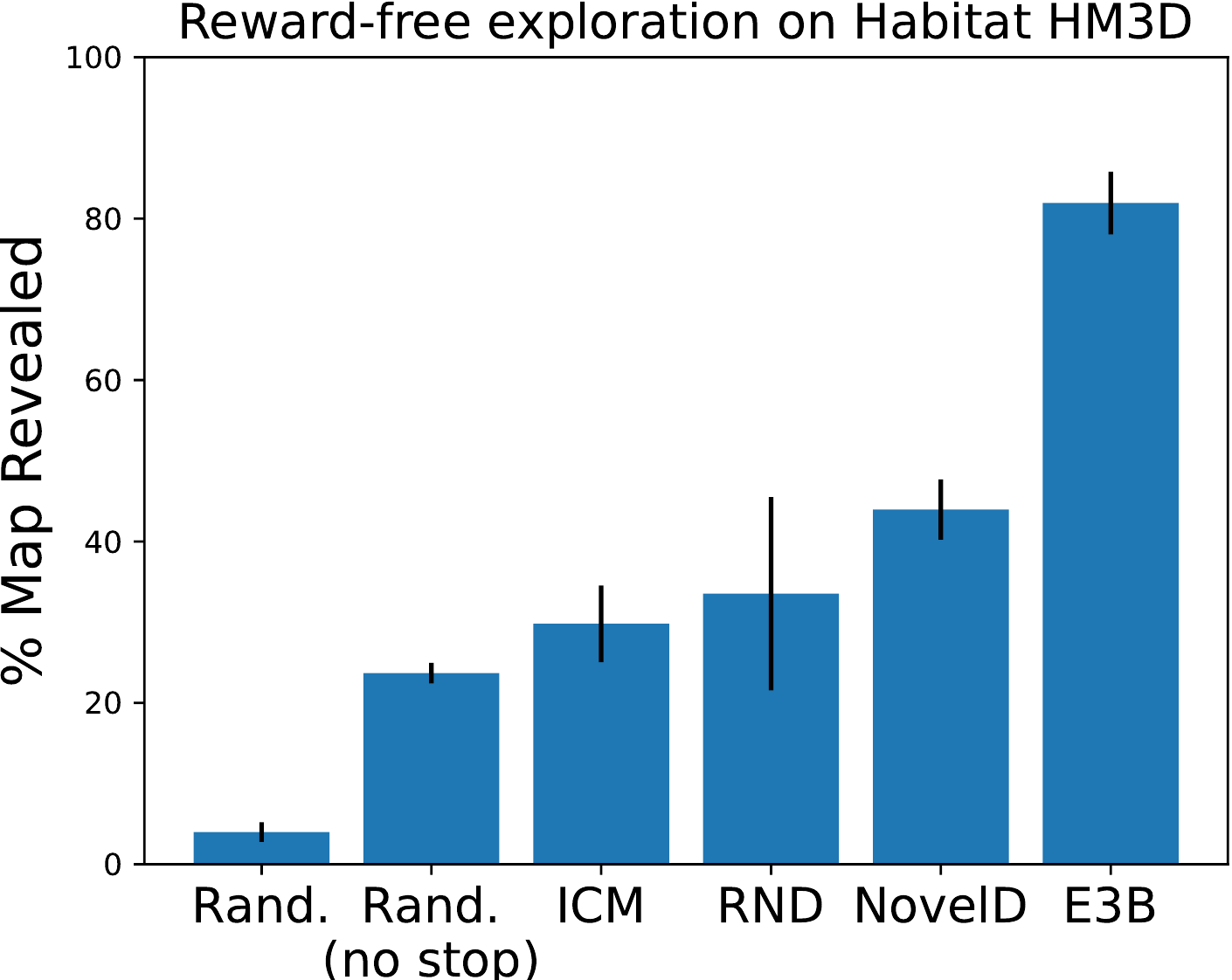}
   \end{center}
   \caption{Reward-free exploration on Habitat. Error bars represent std. deviations over $3$ seeds.}
   \vspace{-6mm}
   \label{fig:habitat}
 \end{wrapfigure}
 Habitat is a platform for embodied AI research which provides an interface for agents to navigate and act in photorealistic simulations of real indoor environments. 
At each episode during training, the agent is initialized in a different environment, and it is tested on a set of held-out environments not used during training. 
These experiments are designed to evaluate exploration of realistic CMDPs with visually rich observations.

We use the HM3D dataset \cite{hm3d}, which contains high-quality renditions of $1000$ different indoor spaces.  %The action space consists of $4$ actions: \texttt{\{move\_forward, turn\_left, turn\_right, stop\_episode\}}.
 As our base RL algorithm we use DD-PPO \cite{DDPPO} and train ICM, RND, NovelD and \alg~agents using the intrinsic reward alone. We then evaluate each agent (as well as two random agents) on unseen test environments by measuring how much of each environment has been revealed by the agent's line of sight over the course of the episode. Full details on the experimental setup can be found in Appendix \ref{appendix:habitat}.

\begin{figure}[h]
     \centering
     \begin{subfigure}[b]{0.23\textwidth}
         \centering
         \fbox{\includegraphics[width=\textwidth, height=0.7\textwidth]{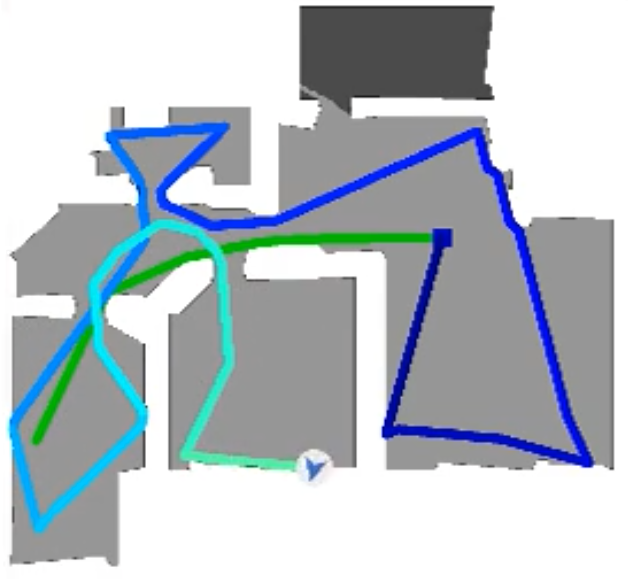}}
         \caption{E3B}
     \end{subfigure}
     \hfill
     \begin{subfigure}[b]{0.23\textwidth}
         \centering
         \fbox{\includegraphics[width=\textwidth, height=0.7\textwidth]{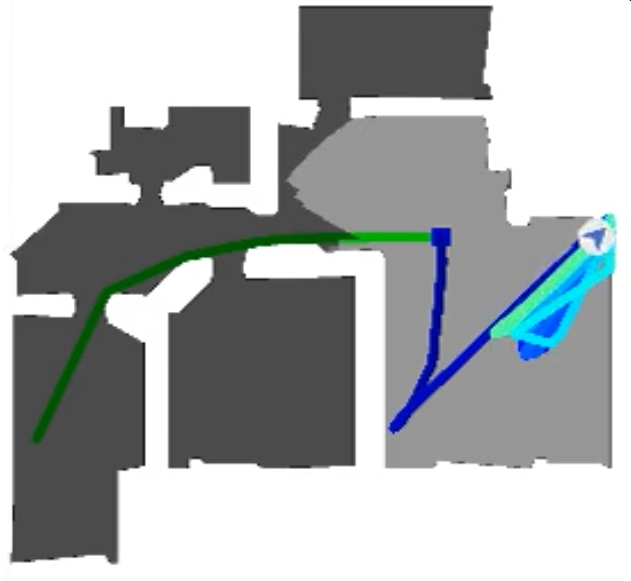}}
         \caption{ICM}
     \end{subfigure}
     \hfill
     \begin{subfigure}[b]{0.23\textwidth}
         \centering
         \fbox{\includegraphics[width=\textwidth, height=0.7\textwidth]{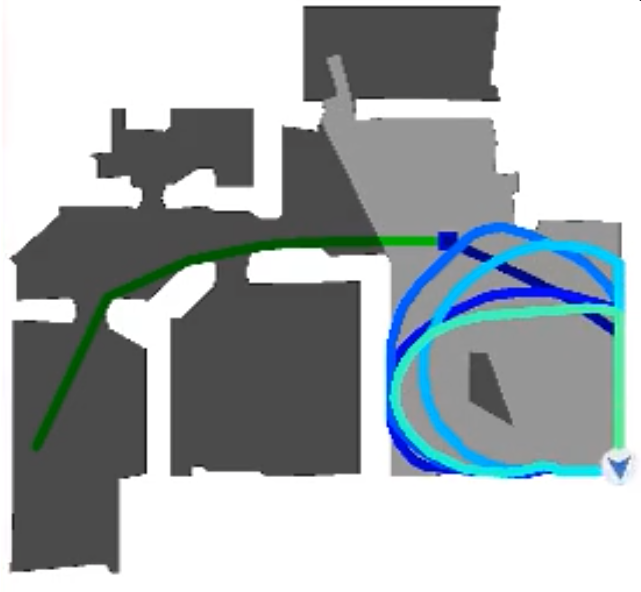}}
         \caption{NovelD}
     \end{subfigure}
     \hfill
     \begin{subfigure}[b]{0.23\textwidth}
         \centering
         \fbox{\includegraphics[width=\textwidth, height=0.7\textwidth]{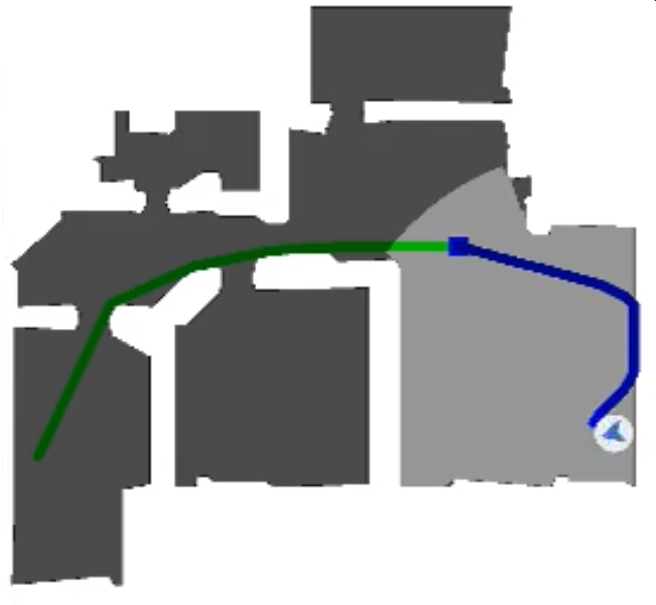}}
         \caption{RND}
     \end{subfigure} 
     \caption{Trajectories of policies trained with different exploration algorithms, on a Habitat environment unseen during training. \alg~ reveals a larger portion of the map than other methods.}
     \label{fig:habitat-trajs}
\end{figure}

Quantitative results are shown in Figure \ref{fig:habitat}. The \alg~agent reveals significantly more of the test maps than any of the other agents. Trajectories for \alg, ICM, RND and NovelD on one of the test maps are shown in Figure \ref{fig:habitat-trajs}, which illustrate how \alg~explores a large portion of the space whereas the others do not. These results provide evidence for \alg's scalability to high-dimensional pixel-based observations, and reinforce its broad applicability. %We also provide results for the pixel-based Vizdoom environment in Appendix B.
% We hypothesize that this is because the global bonuses of RND and NovelD, which measure novelty across all episodes, are poorly suited to this setting where each episode corresponds to a different environment, and comparing observations across episodes is not meaningful. Although NovelD also has an episodic bonus, since it is count-based it does not provide useful signal for high-dimensional observation spaces such as the ones here. In contrast, \alg~explores effectively thanks to the combination of an episodic bonus which operates in continuous spaces, and a feature learning method which maps observations to a compact embedding space. 
%We hypothesize that RND does not obtain good performance because its reward bonus, which measures novelty across all episodes, is ill-suited to the CMDP framework where each episode corresponds to a different environment, and comparing observations across episodes is not meaningful. Although NovelD has an episodic term in its bonus, this is count-based and does not provide meaningful signal for high-dimensional observation spaces such as this. \alg~uses an episodic bonus which applies in high-dimensional continuous observation spaces, which enables it to succeed here. 

\section{Related Work}
\label{sec:related}

\textbf{Exploration in RL.} 
Exploration remains a long-standing problem in RL. Common approaches include $\epsilon$-greedy~\cite{sutton&barto}, count-based exploration~\cite{strehl2008analysis, bellemare2016unifying, ostrovski2017count, martin2017count, tang2017exploration, machado2020count}, curiosity-based exploration~\cite{schmidhuber1991possibility, CS, DeepCS, pathak18largescale}, and other types of intrinsic motivation~\cite{oudeyer2007intrinsic, oudeyer2009intrinsic, stadie2015incentivizing, achiam2017surprise, burda2018exploration}. These methods were largely designed for singleton MDPs, where the environment remains the same across episodes. As a result, they measure novelty over the agent's lifetime in a static environment rather than on an episodic basis under a distribution of diverse contexts, as necessary in CMDPs. Other intrinsic motivation methods have recently been developed for exploration in CMDPs \citep{RIDE, zha2021rank, AGAC, NovelD}. As our experiments show, they critically rely on episodic count-based bonuses, which \alg{} generalizes.

Another class of methods automatically generate curricula over variations of the CMDP to encourage efficient learning, effectively performing a form of curiosity-driven exploration in the context space. These include goal-conditioned ~\cite{forestier2017intrinsically, florensa2017reverse, Fang2019CurriculumguidedHE, Racanire2019AutomatedCT, Colas2020IntrinsicallyMG, campero2020learning, goexplore} and goal-free variants \cite{Sukhbaatar2018IntrinsicMA, Portelas2019TeacherAF, jiang2021prioritized, dennis2020emergent}.
%Goal-conditioned automatic curriculum methods provide the goal subset of the context to the agent ~\cite{forestier2017intrinsically, florensa2017reverse, Fang2019CurriculumguidedHE, Racanire2019AutomatedCT, Colas2020IntrinsicallyMG, campero2020learning, goexplore}, while goal-free variants can explore contexts without \emph{a priori} disclosing any knowledge about the context \cite{Sukhbaatar2018IntrinsicMA, Portelas2019TeacherAF, jiang2021prioritized, dennis2020emergent}. 
Unlike our method, these methods assume the ability to actively configure the environment context. In principle, \alg{} can be combined with these approaches, whereby \alg{} explores at an episodic level, while the automatic curriculum method explores at a context level, making the two complementary.%, making our method complementary to this body of work.

\textbf{Elliptical Bonuses.} The use of elliptical bonuses has a long history in the contextual bandit literature \cite{Auer2002, Dani2008, LinUCB}, which corresponds to the RL setting with a single time step. 
%The LinUCB \cite{LinUCB} algorithm trains a linear regressor for each action, designed to predict the reward and associated confidence interval corresponding to an input context. The action with the highest upper confidence bound is then chosen, which can be viewed as using the environment reward function augmented with an elliptical bonus.
More recently, several works have begun to explore the use of elliptical bonuses in the context of multi-step RL. The PC-PG algorithm \cite{PCPG} considers MDPs with linear dynamics and uses an elliptical bonus based on a policy cover to explore in a provably efficient manner. The ACB algorithm \cite{ACB} also uses an elliptical bonus, approximated using linear regressors trained on random noise, while FLAMBE \cite{FLAMBE} uses an elliptical bonus inside a learned dynamics model. All of these algorithms operate on singleton MDPs, whereas ours is designed for contextual MDPs and constructs elliptical bonuses at the episode level rather than across episodes. We also use different feature learning methods, namely inverse dynamics models, instead of the policy encoders, random networks or kernel methods used in the aforementioned works.

\section{Conclusion}

In this work, we identified a fundamental limitation of existing methods for exploration in CMDPs: their performance relies heavily on an episodic count-based term, which is not meaningful when each state is unique. This is a common scenario in realistic applications, and it is difficult to alleviate through feature engineering.
%While feature engineering can alleviate the issue in some cases, it it difficult to design features which are broadly effective across tasks.
%which is highly dependant on the representation of the observations that is used for counting. While different feature extraction approaches can be useful for certain tasks, it is challenging to design one which is broadly effective. 
%In addition, such count-based methods are limited to tabular state spaces or low-dimensional observations, making them unfeasible for many real-world applications. 
To remedy this limitation, we introduce a new method, \alg, which extends episodic count-based methods to continuous state spaces using an elliptical episodic bonus, as well as an inverse dynamics model to automatically extract useful features from states. \alg~achieves a new state-of-the-art on a wide range of complex tasks from the MiniHack suite, without the need for feature engineering. 
Our approach also scales to high-dimensional pixel-based environments, demonstrated by the fact that it matches top exploration methods on VizDoom and outperforms them in reward-free exploration on Habitat. Future research directions include experimenting with more advanced feature learning methods, and investigating ways to integrate within-episode and across-episode novelty bonuses.
%Our approach also scales to high-dimensional pixel-based environments such as VizDoom, matching the results of other top exploration methods. Future research directions include experimenting with more advanced feature learning methods, and investigating ways to integrate within-episode and across-episode novelty bonuses. 

%While our paper mainly focuses on effective exploration within each episode, an interesting avenue for future work is to better integrate within-episode and across-episodes exploration to maximize information transfer.  %In conclusion, E2B opens the door to exploration methods that work in complex, rich, noisy, and stochastic environments, where the agent rarely sees the same observation more than once. We expect that combining E2B with better representation learning methods (e.g. using contrastive or self-supervised training) could further improve exploration in complex environments. While our paper mainly focuses on effective exploration within each episode, an interesting avenue for future work is to better integrate within-episode and across-episodes exploration to maximize information transfer. 

\medskip

\bibliographystyle{plain}

\bibliography{ref}

\clearpage
%%%%%%%%%%%%%%%%%%%%%%%%%%%%%%%%%%%%%%%%%%%%%%%%%%%%%%%%%%%%
\section*{Checklist}

\begin{enumerate}

\item For all authors...
\begin{enumerate}
  \item Do the main claims made in the abstract and introduction accurately reflect the paper's contributions and scope?
    \answerYes{}{}
  \item Did you describe the limitations of your work?
    \answerYes{} We discuss how our method (as well as existing methods) are not able to solve MiniHack tasks with large action spaces in Section \ref{sec:experiments} as well as Appendix \ref{appendix:action_restriction}. 
  \item Did you discuss any potential negative societal impacts of your work?
    \answerYes{See Appendix \ref{appendix:impact_statement}.}
  \item Have you read the ethics review guidelines and ensured that your paper conforms to them?
    \answerYes{}
\end{enumerate}

\item If you are including theoretical results...
\begin{enumerate}
  \item Did you state the full set of assumptions of all theoretical results?
    \answerNA{}
	\item Did you include complete proofs of all theoretical results?
    \answerNA{}{}
\end{enumerate}

\item If you ran experiments...
\begin{enumerate}
  \item Did you include the code, data, and instructions needed to reproduce the main experimental results (either in the supplemental material or as a URL)?
    \answerYes{}{We included the code with instructions in the supplement, and will release the code on GitHub upon acceptance.}
  \item Did you specify all the training details (e.g., data splits, hyperparameters, how they were chosen)?
    \answerYes{}{See Appendix \ref{appendix:experiments}.}
	\item Did you report error bars (e.g., with respect to the random seed after running experiments multiple times)?
    \answerYes{}{We use the \texttt{rliable} library to compute error bars using stratified bootstrapping in Figure \ref{fig:aggregate}.}
	\item Did you include the total amount of compute and the type of resources used (e.g., type of GPUs, internal cluster, or cloud provider)?
    \answerYes{}{See Appendix \ref{appendix:compute_details}.}
\end{enumerate}

\item If you are using existing assets (e.g., code, data, models) or curating/releasing new assets...
\begin{enumerate}
  \item If your work uses existing assets, did you cite the creators?
    \answerYes{}{We cited all the codebases we built upon, see Section \ref{appendix:codebase_details}.}
  \item Did you mention the license of the assets?
    \answerYes{}{We mentioned all licenses associated with the codebases we used, see Section \ref{appendix:codebase_details}.}
  \item Did you include any new assets either in the supplemental material or as a URL?
    \answerYes{}{All of our new code is included in our code release.}
  \item Did you discuss whether and how consent was obtained from people whose data you're using/curating?
    \answerNA{}{}
  \item Did you discuss whether the data you are using/curating contains personally identifiable information or offensive content?
    \answerNA{}{}
\end{enumerate}

\item If you used crowdsourcing or conducted research with human subjects...
\begin{enumerate}
  \item Did you include the full text of instructions given to participants and screenshots, if applicable?
    \answerNA{}{}
  \item Did you describe any potential participant risks, with links to Institutional Review Board (IRB) approvals, if applicable?
    \answerNA{}{}
  \item Did you include the estimated hourly wage paid to participants and the total amount spent on participant compensation?
    \answerNA{}{}
\end{enumerate}

\end{enumerate}

%%%%%%%%%%%%%%%%%%%%%%%%%%%%%%%%%%%%%%%%%%%%%%%%%%%%%%%%%%%%

\clearpage
\appendix

\section{Broader Impact Statement}
\label{appendix:impact_statement}

This work proposes an RL exploration method for contextual MDPs, which is a very broad framework. Many decision-making problems can be framed as contextual MDPs, such as autonomous driving (contexts represent cars/roads), household robotics (contexts represent houses), healthcare applications (contexts represent patients) and online recommendation/ad optimization (contexts represent customers). Like other RL exploration algorithms, our method facilitates learning a policy which maximizes some reward function specified by a designer. Depending on the goals of the reward function designer, executing the resulting policy could result in positive or negative consequences.

\section{Algorithm Details}
\label{appendix:alg_details}

\begin{algorithm}[h!]
\caption{Exploration via Episodic Elliptical Bonuses (\alg)}
\begin{algorithmic} 
%\REQUIRE $n \geq 0 \vee x \neq 0$
\STATE Initialize policy $\pi$, feature encoder $\phi$ and inverse dynamics model $f$.
\WHILE{not converged}
\STATE Sample context $c \sim \mu_C$ and initial state $s_0 \sim \mu_S(\cdot | c)$
\STATE Initialize inverse covariance matrix: $C_0^{-1} = \frac{1}{\lambda} I$
\FOR{$t = 0, ..., T$}
\STATE $a_t \sim \pi(\cdot | s_t)$  \COMMENT{Sample action}
\STATE $s_{t+1}, r_{t+1} \sim P(\cdot | s_t, a_t)$  \COMMENT{Step through environment}
\STATE $b_{t+1} = \phi(s_{t+1})^\top C_t^{-1} \phi(s_{t+1})$ \COMMENT{Compute bonus}
\STATE $u = C_t^{-1}\phi(s_{t+1})$
\STATE $C_{t+1}^{-1} = C_t^{-1} - \frac{1}{1 + b_{t+1}} uu^\top$ \COMMENT{Update inverse covariance matrix}
\STATE $\bar{r}_{t+1} = r_{t+1} + \beta b_{t+1}$
\ENDFOR
\STATE Perform policy gradient update on $\pi$ using rewards $\bar{r}_1, ..., \bar{r}_T$.
\STATE Update $\phi$ and $g$ using $\{(s_t, a_t, s_{t+1})\}_{t=0}^{T-1}$ to minimize the loss: 
\begin{equation*}
\ell = -\log(p(a_t | f(\phi(s_t), \phi(s_{t+1}))))
\end{equation*}
%$\ell = -\log(p_f(a_t | \phi(s_t), \phi(s_{t+1})))$
\ENDWHILE
\end{algorithmic}
\end{algorithm}

The full algorithm details are shown above. 

\textbf{Additional intuition:} The elliptical bonus is related to the Mahalanobis distance \cite{mahalanobis1936generalized} which uses a similar bilinear form. However, the Mahalanobis distance would normalize the matrix $C_{t-1}$ in equation \ref{eq:elliptical_bonus} by the number of observations $t-1$ in the episode, whereas the elliptical bonus does not.The elliptical bonus thus tends to decrease with the number of observations, similarly to the count-based bonus. 
 \begin{figure}[h]
 \centering
     \includegraphics[width=0.35\textwidth]{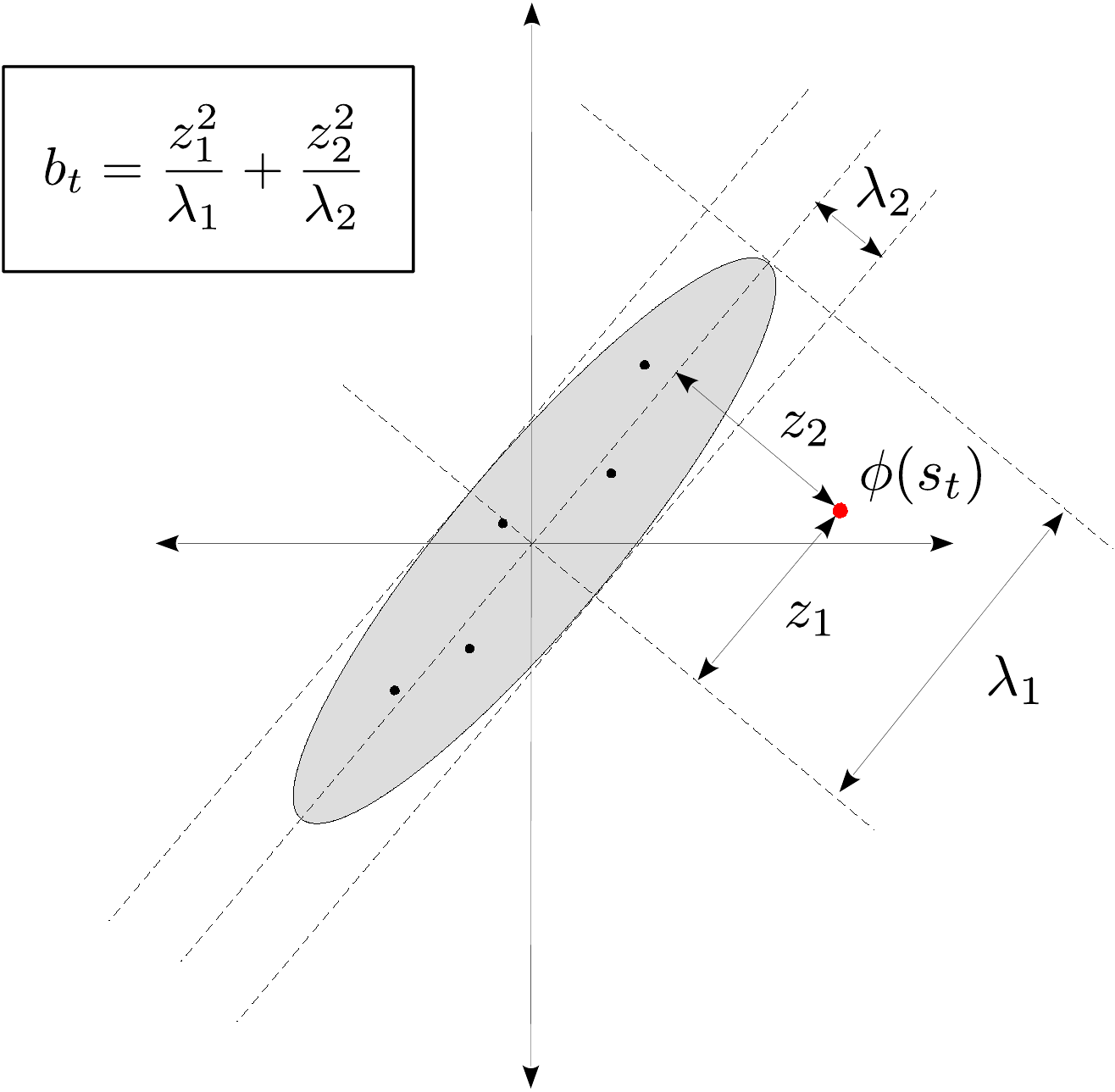}
   \caption{Illustration of the elliptical bonus in 2 dimensions.}
   \label{fig:elliptical2d}
   \vspace{-1mm}
 \end{figure}

\section{Experiment Details}
\label{appendix:experiments}

\subsection{MiniHack}
\label{appendix:minihack}
\subsubsection{Architecture Details}

We follow the policy network architecture described in \cite{minihack}. The policy network has four trunks: i) a 5-layer convolutional trunk which maps the full symbol image (of size $79 \times 21$) to a hidden representation, ii) a second 5-layer convolutional trunk which maps a $9 \times 9$ crop centered at the agent to a hidden representation, iii) an MLP trunk which maps the stats vector to a hidden representation, and iv) a 1-D convolutional trunk with interleaved max-pooling layers, followed by a fully-connected network which maps the message to a hidden representation. The hidden representations are then concatenated together, passed through a 2-layer fully-connected network followed by an LSTM \cite{LSTM} layer. The output of the LSTM layer is then passed to linear layers which produce action probabilities and a value function estimate. 

The convolutional trunks i) and ii) have the following hyperparameters: 5 layers, filter size 3, symbol embedding dimension 64, stride 1, filter number 16 at each layer except the last, which is 8, and ELU non-linearities \cite{ELU}. The MLP trunk iii) has 2 hidden layers of 64 hidden units each with ReLU non-linearities.
The trunk iv) for processing messages has 6 convolutional layers, each with 64 input and output feature maps. The first two have kernel size 7 and the rest have kernel size 3. All have stride 1 and there are max-pooling layers (kernel size 3, stride 3) after the 1st, 2nd and 6th convolutional layers. The last two layers are fully-connected and have 128 hidden units and ReLU non-linearities.

For E3B, we used the same architecture as the policy encoder for the feature embedding $\phi$, except we removed the last layers mapping the hidden representation to the actions and value estimate. The inverse dynamics model is a single-layer fully-connected network with 256 hidden units, mapping two concatenated $\phi$ outputs to a softmax distribution over actions.

\subsubsection{RL Hyperparameters}

For all algorithms we use IMPALA \cite{IMPALA} as our base policy optimizer. Hyperparameters which are common to all methods are shown in Table \ref{tab:hyperparams-common}. All algorithms were trained for 50 million environment steps. We did not anneal learning rates for any of the methods during training, since we found this yielded similar or better performance and simplified the setup. 

Hyperparameters specific to \alg, NovelD, RIDE and ICM are shown in Tables \ref{tab:hyperparams-e3b}, \ref{tab:hyperparams-noveld}, \ref{tab:hyperparams-ride} and \ref{tab:hyperparams-icm}. For both \alg~and NovelD, we experimented with a rolling normalization of the intrinsic reward similar to that proposed in the RND paper \cite{RND}. Specifically, we maintained a running standard deviation $\sigma$ of the intrinsic rewards and divided the intrinsic rewards by $\sigma$ before feeding them to the policy optimizer. We found that this led to improved aggregate performance across environments for \alg~and NovelD (the improvement for NovelD was relatively minor). We found that tuning the intrinsic reward coefficient was important for best performance for all methods, as well as the regularization coefficient of the $C_t$ matrix $\lambda$ for \alg. For RIDE, we used the default hyperparameters from the RIDE implementation in the MiniHack paper \cite{minihack} and otherwise tuned the intrinsic reward coefficient. For ICM, we used the same hyperparameters for the forward and inverse models as for RIDE and also tuned the intrinsic reward coefficient.

\begin{table}[h]
  \caption{Common IMPALA Hyperparameters for MiniHack}
  \centering
  \begin{tabular}{ll}
    \toprule
    \midrule
    Learning Rate & $0.0001$  \\
    RMSProp smoothing constant & $0.99$ \\
    RMSProp momentum & $0$ \\
    RMSProp $\epsilon$ & $10^{-5}$ \\
    Unroll Length     & $80$ \\
    Number of buffers     & $80$       \\
    Number of learner threads & $4$ \\
    Number of actor threads & $256$ \\
    Max gradient norm & $40$ \\
    Entropy Cost & $0.0005$ \\
    Baseline Cost & $0.5$ \\
    Discounting Factor & $0.99$ \\
    \bottomrule
  \end{tabular}
  \label{tab:hyperparams-common}
\end{table}

\begin{table}[h!]
  \caption{Hyperparameters for \alg}
  \vspace{1mm}
  \centering
  \begin{tabular}{|l|l|l|}
    \hline
    Hyperparameter & Values considered & Final Value \\
    \hline
    Running intrinsic reward normalization & $\{$\texttt{True, False}$\}$ & \texttt{True} \\
    Ridge regularizer $\lambda$ & $\{1.0, 0.1, 0.01\}$ & $0.1$  \\
    Entropy Cost & $\{0.0005, 0.005\}$  & $0.005$ \\    
    Intrinsic reward coefficient $\beta$ & $\{0.0001, 0.001, 0.01, 0.1, 1, 10\}$  & $1$ \\
    \hline
  \end{tabular}
  \label{tab:hyperparams-e3b}
  \vspace{2mm}
  \caption{Hyperparameters for NovelD}
  \centering
  \begin{tabular}{|l|l|l|}
    \hline
    Hyperparameter & Values considered & Final Value \\
    \hline
    Running intrinsic reward normalization & $\{$\texttt{True, False}$\}$ & \texttt{True}  \\
    Scaling factor $\alpha$ & $\{0.1, 0.5\}$ & $0.1$ \\
    Entropy Cost & $\{0.0005, 0.005\}$  & $0.005$ \\     
    Intrinsic reward coefficient $\beta$ & $\{0.001, 0.01, 0.1, 1, 10, 100\}$  & $1$ \\
    \hline
  \end{tabular} 
    \label{tab:hyperparams-noveld}
  \vspace{2mm}
  \caption{Hyperparameters for RIDE}
  \centering
  \begin{tabular}{|l|l|l|}
    \hline
    Hyperparameter & Values considered & Final Value \\
    \hline
    Forward Model loss coefficient & $1.0$ & $1.0$ \\
    Inverse Model loss coefficient & $0.1$ & $0.1$ \\
    Entropy Cost & $\{0.0005, 0.005\}$  & $0.0005$ \\     
    Intrinsic reward coefficient $\beta$ & $\{0.001, 0.01, 0.1, 1, 10, 100\}$  & $0.1$ \\
    \hline
  \end{tabular}
    \label{tab:hyperparams-ride}
  \vspace{2mm}
  \caption{Hyperparameters for ICM}
  \centering
  \begin{tabular}{|l|l|l|}
    \hline
    Hyperparameter & Values considered & Final Value \\
    \hline
    Forward Model loss coefficient & $1.0$ & $1.0$ \\
    Inverse Model loss coefficient & $0.1$ & $0.1$ \\
    Entropy Cost & $\{0.0005, 0.005\}$  & $0.0005$ \\     
    Intrinsic reward coefficient $\beta$ & $\{0.001, 0.01, 0.1, 1, 10, 100\}$  & $0.1$ \\
    \hline
  \end{tabular}  
  \label{tab:hyperparams-icm}
  
\end{table}

\clearpage

\subsubsection{Environment Details}
\label{appendix:env_details}

We used $16$ MiniHack environments in total. These include the following $9$ navigation-based tasks: 

\texttt{
         'MiniHack-MultiRoom-N4-Locked-v0',
         'MiniHack-MultiRoom-N6-Lava-v0',
         'MiniHack-MultiRoom-N6-Lava-OpenDoor-v0',
         'MiniHack-MultiRoom-N6-LavaMonsters-v0',
         'MiniHack-MultiRoom-N10-OpenDoor-v0',
         'MiniHack-MultiRoom-N10-Lava-OpenDoor-v0',
         'MiniHack-LavaCrossingS19N13-v0',
         'MiniHack-LavaCrossingS19N17-v0',
         'MiniHack-Labyrinth-Big-v0'
}

as well as the following $7$ skill-based tasks: 

\texttt{'MiniHack-Levitate-Potion-Restricted-v0',
         'MiniHack-Levitate-Boots-Restricted-v0',
         'MiniHack-Freeze-Horn-Restricted-v0',
         'MiniHack-Freeze-Wand-Restricted-v0',
         'MiniHack-Freeze-Random-Restricted-v0',
         'MiniHack-LavaCross-Restricted-v0',
         'MiniHack-WoD-Hard-Restricted-v0'}

The \texttt{MultiRoom-N4-Locked} and \texttt{MultiRoom-N*-Lava}, \texttt{Labyrinth-Big}, \texttt{LavaCrossingS*N*}, as well as all the skill-based tasks, are taken from the official MiniHack github repository (\url{https://github.com/facebookresearch/minihack}). We made some of these harder by increasing the number of rooms.  

We also include some new environments which we designed to better test the limits of the different algorithms. Specifically, \texttt{'MiniHack-MultiRoom-N*OpenDoor-v0'} are variants on the standard \texttt{MultiRoom} tasks, the only difference being that the doors connecting the rooms are initialized to be open rather than closed. This is because opening a door causes a message to appear "the door opens". By initializing the door to be closed, the agent does not see any messages when passing from one room to the next. As discussed in Appendix \ref{appendix:additional_discussion}, this seemingly trivial change can cause the NovelD-message variant to fail completely, shedding light on its lack of robustness.

We found that the \texttt{MiniHack-MultiRoom-N6-Extreme} task was impossible to solve consistently even for a human player due to the large quantities of monsters, so we did not include it. We instead included a variant with less monsters and lava called \texttt{MultiRoom-N6-LavaMonsters}. 

\subsubsection{Action Space Restriction}
\label{appendix:action_restriction}

In their default setup, the skill-based tasks have a large context-dependent action space of $78$ actions. Many of these actions are unrelated to the task at hand, but produce unique in-game messages when executed which nevertheless do not affect the underlying state of the MDP. For example, trying to pay a non-existent shopkeeper by executing the PAY action results in the message "There appears to be no shopkeeper here to receive your payment.", and executing the FIGHT action in the absence of adversaries results in the message "You attack thin air." We found that neither the baselines nor our proposed method were able to solve the skill-based tasks with the full action space. For the NovelD variants which use positions or symbolic images for the episodic counts, this is reasonable since the tasks are not navigation-based (these methods did not work even with restricted action spaces). For NovelD with the message-based bonus, we found that the agent ends up learning a policy where it executes many different actions which produce different messages, but do not change the underlying state of the game (such as the PAY and FIGHT actions described above). This makes sense from the perspective of optimizing intrinsic reward, since each new message seen within the episode provides additional intrinsic reward; however, this does not help explore the state space in a way that is helpful for discovering the true environment reward. We observed similar behavior for \alg, and hypothesize that the encoding learned through the inverse dynamics model keeps message information since this is useful for predicting actions, even if they have no real effect (such as PAY and FIGHT). In this case, the policy can then maximize intrinsic reward by executing these actions. 

To avoid this unwanted behavior (which applies to all the methods we tested), we restricted the action space to only the actions which were useful for the tasks at hand. These included actions corresponding to direction movements, as well as different combinations of PICKUP, QUAFF, ZAP, FIRE, and WEAR (full details can be found in our code release). We denote the versions of the tasks with action space restriction with the "\texttt{-Restricted-v0}" suffix, as opposed to the original "\texttt{-v0}" suffix. We believe that better understanding this failure mode and designing exploration bonuses and/or representation learning methods which are robust to it is an important direction for future work. 

\subsubsection{Compute Details}
\label{appendix:compute_details}

Each algorithm was trained using $40$ Intel(R) Xeon(R) CPU cores (E5-2698 v4 @ 2.20GHz) and one NVIDIA GP100 GPU. We used PyTorch \cite{pytorch} for all our experiments. Each run took between approximately $10$ and $30$ hours to complete. The total runtime depended on two factors: the computation time of the algorithm, and the behavior of the policy. In terms of algorithm computation time, \alg~was roughly $1.5$ to $2$ times slower than the other methods, due to the fact that in our implementation an additional forward pass through the embedding network (used to compute the elliptical bonus) was performed on CPU. It is possible that a different implementation where this step would be done on GPU would be faster.

A second factor which influenced the total runtime was how quickly the agent learned to avoid dying. For certain environments (for example, those which contain lava), the agent can die quickly which causes the environment to be regenerated. For environments which call the MiniGrid library on the backend, this can be a performance bottleneck since generating new MiniGrid environments can be slow. We found that algorithms which cause the agent to die frequently were much slower on the \texttt{MiniHack-MultiRoom-N*Lava*} and \texttt{MiniHack-LavaCrossing*} environments, both of which are based on MiniGrid.

\clearpage

\subsection{VizDoom}
\label{appendix:vizdoom_hps}

\subsubsection{Architecture Details}

For all VizDoom experiments, we used the same policy network architecture as in \cite{RIDE}: four 2D convolutional layers with $32$ input and $32$ output channels, kernel size $3 \times 3$, stride $2$ and padding $1$, interleaved with Exponential Linear Unit (ELU) non-linearities \cite{ELU}. The output of the convolutional layers feeds into a 2-layer LSTM \cite{LSTM} with $256$ hidden units, followed by linear layers mapping the hidden units to a distribution over actions and a scalar value estimate. 

The architecture for \alg's encoder is identical to that of the policy network, except that the LSTM is replaced by a linear layer (there is thus no recurrence in the encoder network). 

\subsubsection{RL Hyperparameters}

For our VizDoom experiments, we also used IMPALA as our base policy optimizer. For \alg~we used the same IMPALA hyperparameters as for MiniHack. The hyperparameters specific to \alg, ICM and RIDE are given below. For both ICM and RIDE, we found that performance was quite sensitive to the coefficients of the forward and inverse dynamics model losses. We ended up using the hyperparameters from \cite{RIDE}, which we found gave the best performance.

\begin{table}[h!]
  \caption{Hyperparameters for \alg~on VizDoom}
  \centering
  \begin{tabular}{|l|l|l|}
    \hline
    Hyperparameter & Values considered & Final Value \\
    \hline
    Running intrinsic reward normalization & $\{$\texttt{False}$\}$ & \texttt{False} \\
    Ridge regularizer $\lambda$ & $\{0.1, 0.01\}$ & $0.1$  \\
    Intrinsic reward coefficient $\beta$ & $\{3\cdot10^{-7}, 10^{-6}, 3\cdot10^{-6}, 10^{-5}, 3\cdot10^{-5}, 10^{-4}\}$  & $3\cdot10^{-6}$ \\
    \hline
  \end{tabular}
  \vspace{2mm}
  \caption{Hyperparameters for RIDE on VizDoom}
  \begin{tabular}{|l|l|l|}
    \hline
    Hyperparameter & Values considered & Final Value \\
    \hline
    Running intrinsic reward normalization & $\{$\texttt{False}$\}$ & \texttt{False} \\
    Forward loss coefficient & $\{0.5, 1.0\}$ & $0.5$ \\
    Inverse loss coefficient & $\{0.1, 0.8\}$ & $0.8$\\
    Intrinsic reward coefficient $\beta$ & $\{10^{-2}, 3\cdot10^{-3}, 10^{-3}, 3\cdot10^{-4}, 10^{-4}\}$  & $3\cdot10^{-2}$ \\
    \hline
  \end{tabular}  
  \vspace{1mm}
  \caption{Hyperparameters for ICM on VizDoom}
  \begin{tabular}{|l|l|l|}
    \hline
    Hyperparameter & Values considered & Final Value \\
    \hline
    Running intrinsic reward normalization & $\{$\texttt{False}$\}$ & \texttt{False} \\
    Forward loss coefficient & $\{0.2, 1.0\}$ & $0.2$ \\
    Inverse loss coefficient & $\{0.1, 0.8\}$ & $0.8$ \\
    Entropy cost & $\{0.0001, 0.005\}$ & $0.005$ \\
    Intrinsic reward coefficient $\beta$ & $\{10^{-2}, 3\cdot10^{-3}, 10^{-3}, 3\cdot10^{-4}, 10^{-4}\}$  & $3\cdot10^{-3}$ \\
    \hline
  \end{tabular}  
  \label{tab:hyperparams}
\end{table}

\clearpage

\subsection{Habitat}
\label{appendix:habitat}

\subsubsection{Environment Details}

\begin{figure}[h]
     \centering
     \begin{subfigure}[b]{0.6\textwidth}
         \centering
         \includegraphics[width=\textwidth]{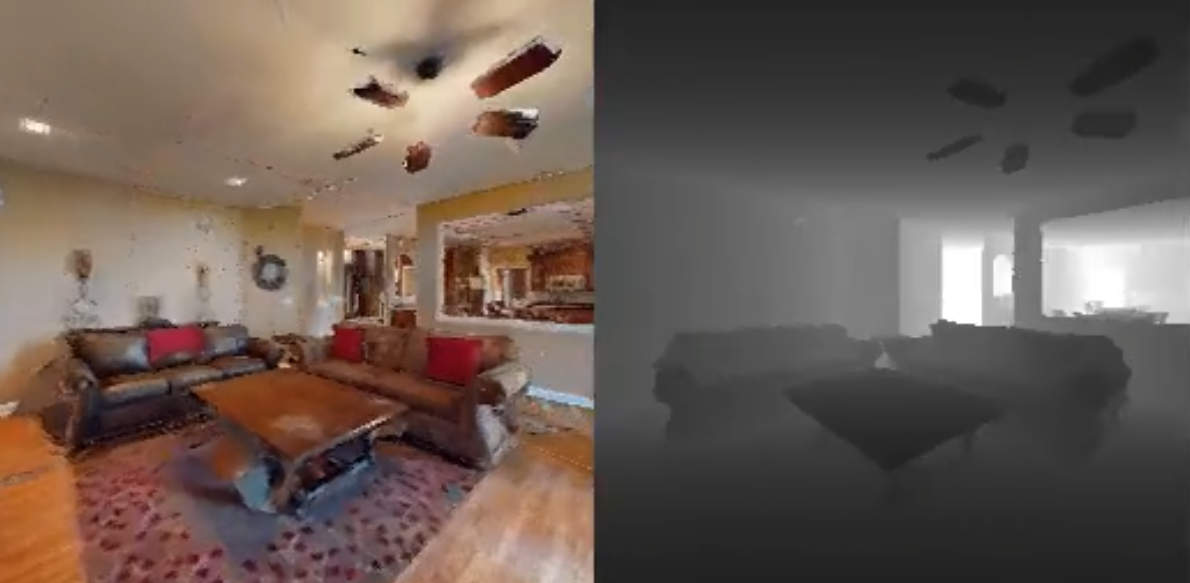}
         \caption{}
     \end{subfigure}
     \hfill
     \begin{subfigure}[b]{0.3\textwidth}
         \centering
         \includegraphics[width=\textwidth]{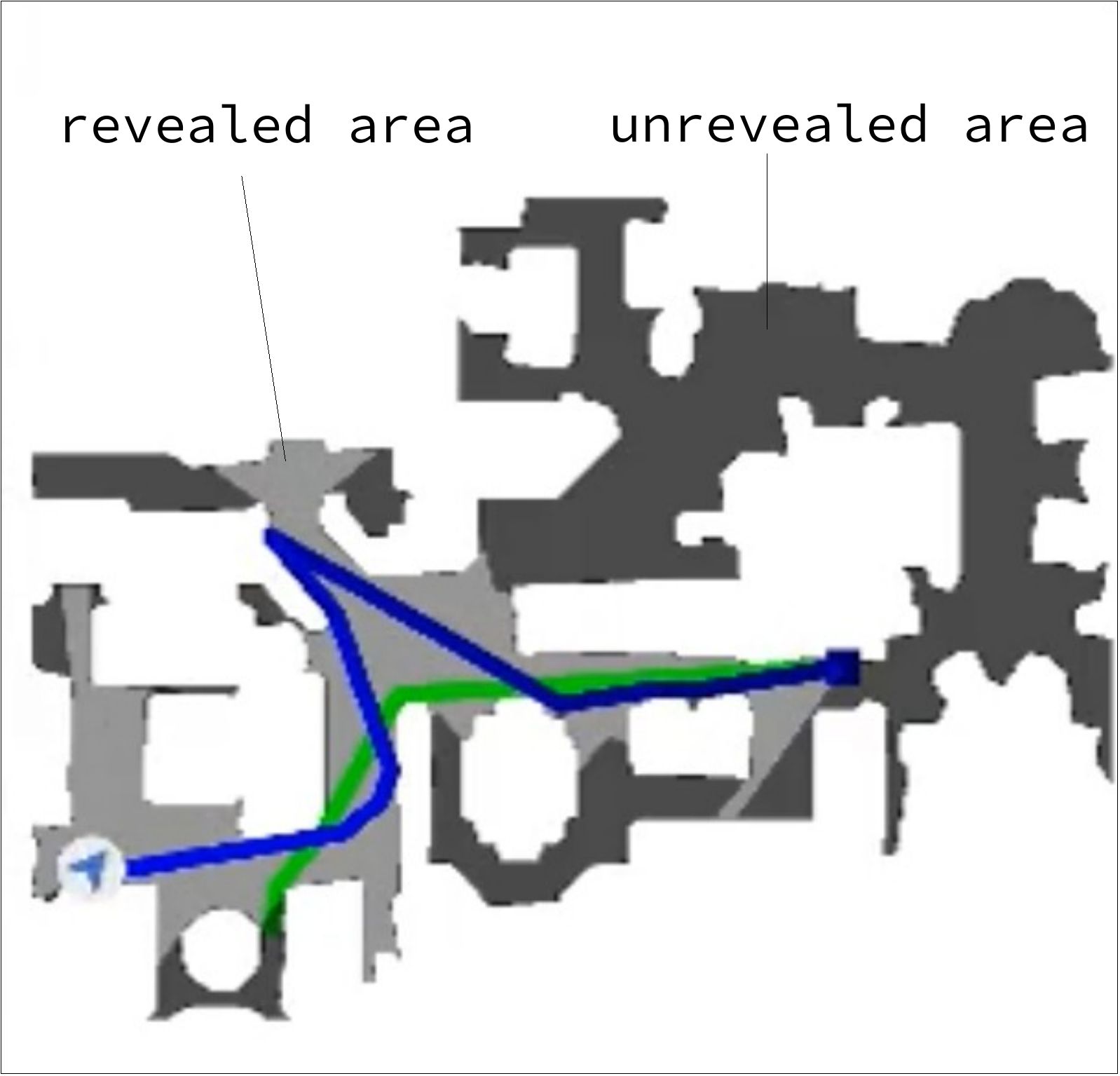}
         \caption{}
     \end{subfigure}
     \caption{a) Visual observations in Habitat  b) Exploration is measured as the proportion of the environment revealed by the agent's line of sight over the course of the episode.}
     \label{fig:habitat2}
\end{figure}

We used the HM3D \cite{hm3d} dataset, which consists of $1000$ high-quality renderings of indoor scenes. 
Observations consist of $4$ modalities: an RGB and depth image (shown in Figure \ref{fig:habitat2}a), GPS coordinates and the compass heading. The action space consists of $4$ actions: $\mathcal{A}$ = \texttt{\{stop\_episode, move\_forward (0.25m), turn\_left ($10^\circ$), turn\_right ($10^\circ$)\}}. The dataset scenes are split into $800/100/100$ train/validation/test splits. Since the test split is not publicly available, we evaluate all models on the validation split. Each scene corresponds to a different context $c \in \mathcal{C}$ in the CMDP framework. 

To measure exploration coverage, we compute the area revealed by the agent's line of site using the function provided by the Habitat codebase \footnote{\url{https://github.com/facebookresearch/habitat-lab/blob/main/habitat/utils/visualizations/fog_of_war.py}}, which uses a modified version of Bresenham's line cover algorithm. We define the exploration coverage to be:

\begin{equation*}
    \mbox{coverage} = \frac{\mbox{revealed area}}{\mbox{total area}}
\end{equation*}

See Figure \ref{fig:habitat2}b) for an illustration. For the results in Figure \ref{fig:habitat}, we evaluated exploration performance for each algorithm by measuring its coverage on $100$ episodes using scenes from the validation set (which were not used for training). 

\subsubsection{Architecture Details}

For all Habitat experiments we used the same policy network as in \cite{DDPPO}, which includes a ResNet50 visual encoder \cite{resnet} and a 2-layer LSTM \cite{LSTM} policy. In addition to RGB and Depth images, the agent also receives GPS coordinates and compass orientation, represented by $3$ scalars total, which are fed into the policy. See the official code release at \url{https://github.com/facebookresearch/habitat-lab/tree/main/habitat_baselines} for full details. 

For exploration algorithms which use inverse dynamics models (\alg~and ICM), we set the architecture of the encoder $\phi$ to be identical to that of the policy network, except that the last layer mapping hidden units to actions is removed. The inverse dynamics model was a single layer MLP with $256$ hidden units and ReLU non-linearities. 

For exploration algorithms which use random network distillation (RND and NovelD), we set the architecture of the random network to be identical to that of the policy network.

\subsubsection{RL Hyperparameters}

The DD-PPO hyperparameters which are common to all the algorithms are listed in Table \ref{tab:hyperparams-common-habitat}. The hyperparameters which are specific to each algorithm are listed in Table \ref{tab:hyperparams-habitat-e3b}, \ref{tab:hyperparams-habitat-rnd}, \ref{tab:hyperparams-noveld}. For NovelD's count-based bonus, hashing the full image was too slow to be practical, so we subsampled images by a factor of $1000$ used that for the count-based bonus, along with the GPS coordinates and compass direction.

\begin{table}[h!]
  \caption{Common PPO/DD-PPO Hyperparameters for Habitat}
  \centering
  \begin{tabular}{ll}
    \toprule
    \midrule
    Clipping & $0.2$  \\    
    PPO epochs & $2$  \\    
    Number of minibatches & $2$ \\
    Value loss coefficient & $0.5$ \\
    Entropy coefficient & 0.00005 \\
    Learning rate & $0.00025$ \\
    $\epsilon$ & $10^{-5}$ \\
    Max gradient norm & $0.2$ \\
    Rollout steps & $128$ \\
    Use GAE & True \\
    $\gamma$ & 0.99 \\
    $\tau$ & 0.95 \\
    Use linear clip decay & False \\
    Use linear LR decay & False \\
    Use normalized advantage & False \\
    Hidden size & $512$ \\
    DD-PPO Sync fraction & $0.6$ \\
    \bottomrule
  \end{tabular}
  \label{tab:hyperparams-common-habitat}
\end{table}

\begin{table}[h!]
  \caption{Hyperparameters for \alg~on Habitat}
  \centering
  \begin{tabular}{|l|l|l|}
    \hline
    Hyperparameter & Values considered & Final Value \\
    \hline
    Ridge regularizer $\lambda$ & $\{0.1\}$ & $0.1$  \\
    Intrinsic reward coefficient $\beta$ & $\{1.0, 0.1, 0.01, 0.001, 0.0001$ & $0.1$ \\
    Inverse Dynamics Model updates per PPO epoch & $3$ & $3$ \\
    \hline
  \end{tabular}
  \label{tab:hyperparams-habitat-e3b}
  \vspace{2mm}
  \caption{Hyperparameters for RND on Habitat}
  \begin{tabular}{|l|l|l|}
    \hline
    Hyperparameter & Values considered & Final Value \\  
  \hline
    Intrinsic reward coefficient $\beta$ & $\{1.0, 0.1, 0.01, 0.001, 0.0001$ & $0.1$ \\
    Predictor Model updates per PPO epoch & $3$ & $3$ \\
    \hline
  \end{tabular}  
  \label{tab:hyperparams-habitat-rnd}
  \vspace{1mm}
  \caption{Hyperparameters for NovelD on Habitat}
  \begin{tabular}{|l|l|l|}
    \hline
    Hyperparameter & Values considered & Final Value \\
    \hline
    Intrinsic reward coefficient $\beta$ & $\{1.0, 0.1, 0.01, 0.001, 0.0001$ & $0.1$ \\
    Predictor Model updates per PPO epoch & $3$ & $3$ \\
    Scaling factor $\alpha$ & $0.1$ & $0.1$ \\
    \hline
  \end{tabular}  
  \label{tab:hyperparams-habitat-noveld}
  \caption{Hyperparameters for ICM on Habitat}
  \begin{tabular}{|l|l|l|}
    \hline
    Hyperparameter & Values considered & Final Value \\
    \hline
    Intrinsic reward coefficient $\beta$ & $\{1.0, 0.1, 0.01, 0.001, 0.0001$ & $0.1$ \\
    Forward Dynamics Model loss coefficient & $\{1.0\}$ & $1.0$ \\
%    Predictor Model updates per PPO epoch & $3$ & $3$ \\
    \hline
  \end{tabular}  
  \label{tab:hyperparams-habitat-icm}

\end{table}

\subsubsection{Compute Details}

Each job was run for $225$ million steps, which took approximately $3$ days on $32$ GPUs with $10$ CPU threads.

\clearpage

\subsection{Codebases Used}
\label{appendix:codebase_details}

Our codebase was built atop the following codebases:

\begin{itemize}
    \item The official NovelD codebase: \url{https://github.com/tianjunz/NovelD} (Creative Commons Attribution-NonCommercial 4.0 license) for NovelD, RND, RIDE and count-based baselines (this codebase is build atop the official RIDE codebase below)
    \item The official MiniHack codebase: \url{https://github.com/facebookresearch/minihack} for network architectures appropriate to MiniHack environments (Apache 2.0 license)
    \item The official RIDE codebase: \url{https://github.com/facebookresearch/impact-driven-exploration} (Creative Commons Attribution-NonCommercial 4.0 license) for network architectures appropriate to VizDoom environments
    \item The official Habitat codebase: \url{https://github.com/facebookresearch/habitat-lab/tree/main/habitat_baselines} for Habitat experiments
\end{itemize}

\section{Additional Results and Discussion}
\label{appendix:additional_discussion}

\subsection{Additional MiniGrid Results}
\label{appendix:additional_minigrid_results}

\begin{figure}[h!]
    \centering
    \includegraphics[width=\textwidth]{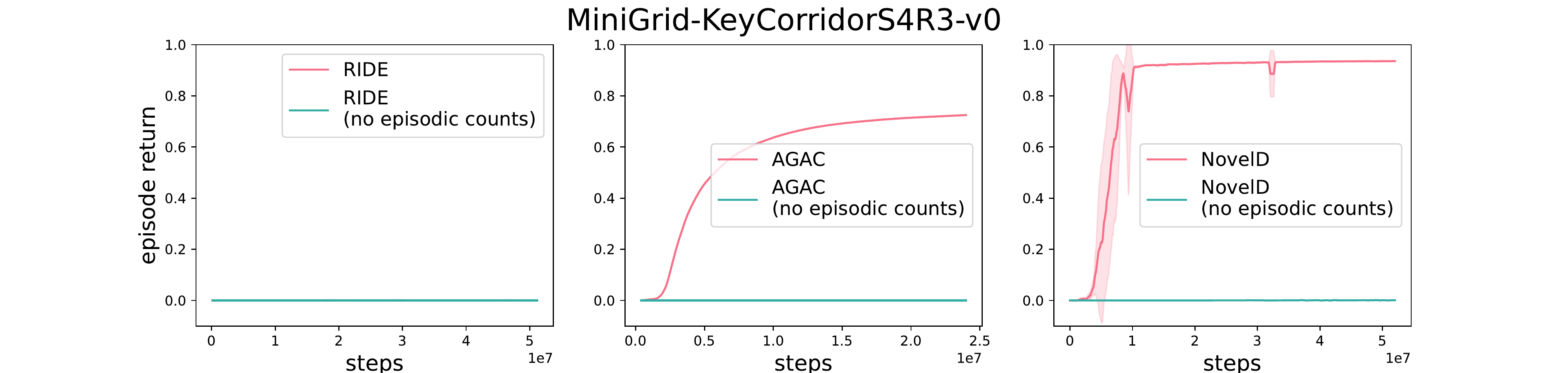} 
    \includegraphics[width=\textwidth]{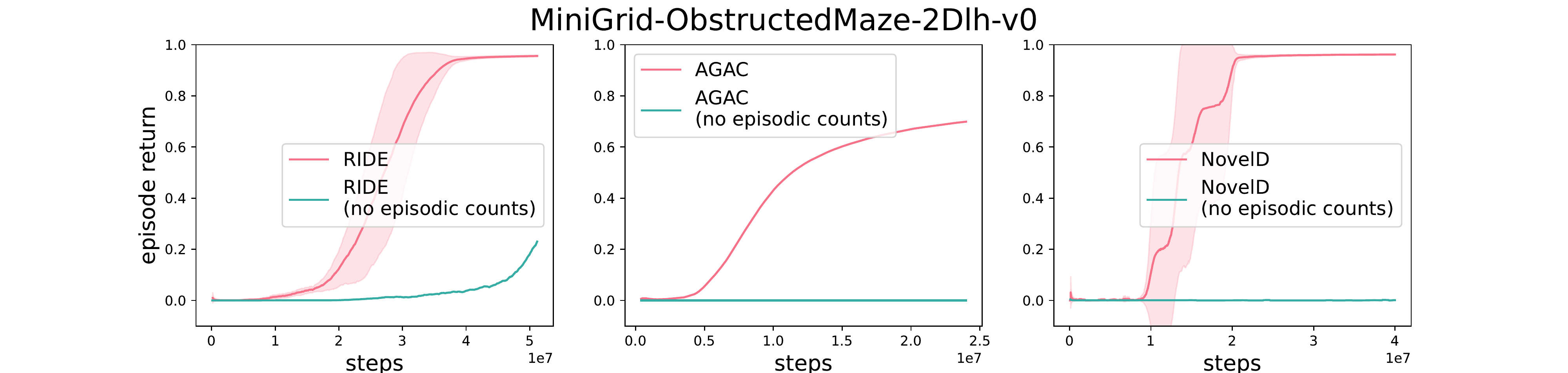}
    \caption{Results for RIDE, AGAC and NovelD with and without the count-based episodic bonus, over $5$ random seeds ($1$ seed for AGAC). Shaded region indicates one standard deviation.}
    \label{fig:counts_ablation2}
\end{figure}

Here we provide results for RIDE, AGAC and NovelD on additional MiniGrid environments used in prior work, with and without their count-based episodic terms. For all algorithms, we used the code released by the authors with the default hyperparameters \footnote{NovelD: \url{https://github.com/tianjunz/NovelD}, RIDE: \url{https://github.com/facebookresearch/impact-driven-exploration}, AGAC: \url{https://github.com/yfletberliac/adversarially-guided-actor-critic}}. Results in Figure \ref{fig:counts_ablation2} confirm the trend in Figure \ref{fig:counts_ablation}: in all cases, removing the count-based episodic term results in a failure to learn or makes learning much slower (such as for RIDE in \texttt{MiniGrid-ObstructedMaze-2Dlh-v0}). We ran less seeds for AGAC because the official release's multithreading implementation was not compatible with our cluster. 

\subsection{Effect of rank-$1$ updates}
\label{appendix:rank1_exp}

Here we compare the wall-clock time for computing the elliptical bonus using the rank-$1$ updates and the naive method of maintaining and inverting the full covariance matrix. We performed these experiments on the \texttt{MiniHack-Freeze-Random-Restricted-v0} environment using the standard hyperparameter setup, on a machine with $80$ Intel(R) Xeon(R) CPU cores (E5-2698 v4 @ 2.20GHz) and one NVIDIA GP100 GPU:

\begin{table}[ht]
    \centering
    \begin{tabular}{c|c}
        \alg~using rank-$1$ update & $767$ FPS  \\
        \hline
        \alg~using matrix inversion & $256$ FPS  \\
    \end{tabular}
\end{table}

This shows that using the rank-$1$ update is essential for fast performance.

\subsection{Negative results for RIDE with modified count-based bonuses}
\label{appendix:ride_modification}

We ran some preliminary experiments investigating modifications to RIDE similar to those for NovelD. Specifically, the modified RIDE reward bonus is:

\begin{equation*}
    b(s_t) = \|\phi(s_{t+1}) - \phi(s_t)\|_2 \cdot \frac{1}{\sqrt{N_e(\psi(s_{t+1}))}}
\end{equation*}

where $\phi$ is the embedding learned with the inverse dynamics model and $\psi(s_{t+1})$ extracts some aspect of the state $s_{t+1}$. We investigated a version where $\psi(s_{t+1})$ extracts the $(x_{t+1}, y_{t+1})$ position information from the state (this method we called \textsc{RIDE-Position} and is the same version that was run in \cite{minihack}) and a version where $\psi$ extracts the message portion of the state (this method we called \textsc{RIDE-Message}). Despite tuning the intrinsic reward coefficient over the range $\{0.0001, 0.001, 0.01, 0.1, 1, 10\}$ on two tasks (\texttt{MiniHack-MultiRoom-N10-v0} and \texttt{MiniHack-Freeze-Horn-Restricted-v0}) over $3$ random seeds, none of the seeds was able to achieve positive reward for either of the tasks with either of the methods. Note that our results are consistent with those reported in \cite{minihack}, which also found that \textsc{RIDE-Position} did not outperform IMPALA on most tasks.

\subsection{Additional MiniHack Results and Discussion}
\label{appendix:additional_minihack_discussion}

% Results for all methods on individual navigation tasks are shown in Figure \ref{fig:nav_task_results} and results on individual skill-based tasks are shown in Figure \ref{fig:skill_task_results}.

\begin{figure}
    \centering
    \includegraphics[width=\textwidth]{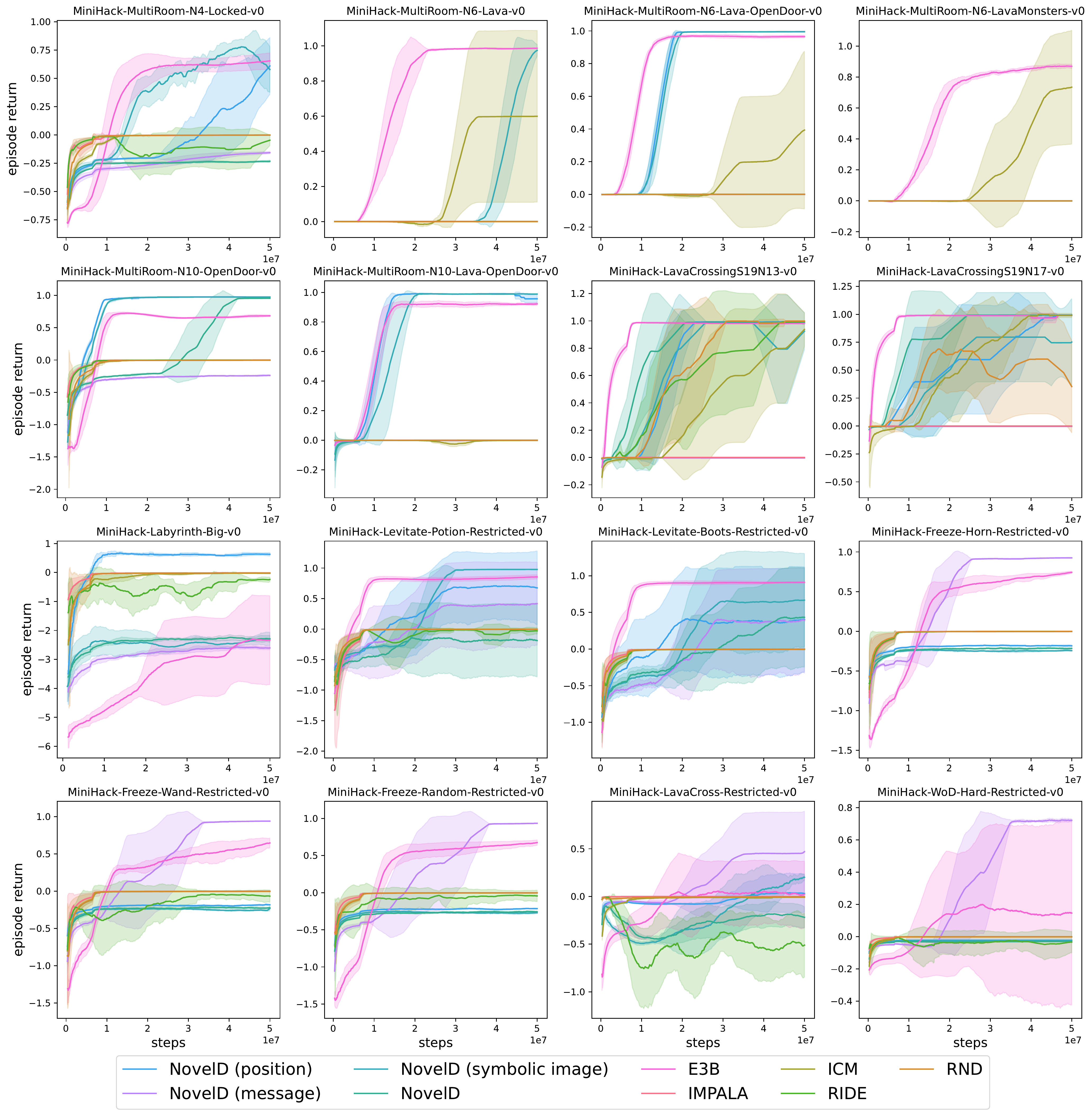}
    \caption{Mean results on individual tasks over $5$ different seeds. Shaded region indicates one standard deviaton.}
    \label{fig:all_big_table}
\end{figure}

 Results for all methods on individual tasks are shown in Figure \ref{fig:all_big_table}. The first $9$ tasks are navigation-based while the remaining $7$ are skill-based.

First, as noted in the main text we see that when using the position bonus, NovelD succeeds in most of the tasks which primarily require the agent to spatially explore the environment, such as the \texttt{Labyrinth-Big} task or the different \texttt{MultiRoom} variants. On the other hand, this modality performs poorly on tasks based on skill acquisition, such as \texttt{Freeze}, \texttt{LavaCross} and \texttt{WoD} tasks. These tasks require the agent to pick up and use objects, and exploring the state space involves trying different action combinations which do not affect the agent's position. Therefore, an exploration bonus which only encourages the agent to visit many different positions is not helpful. 

The message bonus, on the other hand, succeeds on the skill acquisition tasks. On these tasks, the correct sequence of actions indeed causes a sequence of novel messages to appear. For example, let us consider the task \texttt{Freeze-Wand}. In this task, the agent is in a small room and must go to a wand, pick it up, pick the ZAP action, choose the wand and then choose a direction to zap. When conducting this sequence of actions, it encounters the following messages: "you see here a brass wand" (after navigating to the wand location), "f - a brass wand" (after executing the PICKUP action), "What do you want to zap? [f or ?*]" (after executing the ZAP action). Each time the agent visits a state-action pair required to solve the task, it receives one of these messages which provides it with a positive reward signal thanks to the message-based episodic bonus. This in turn reinforces the behavior leading to that state-action pair and allows it to ultimately solve the task.

When using the message bonus, NovelD fails completely on most of the navigation-based tasks, such as \texttt{MultiRoom-N*-Lava-OpenDoor}, \texttt{LavaCrossingS19N13} and \texttt{LavaCrossingS19N17}. For these tasks, the agent must navigate its way through a series of rooms with lava walls or lava rivers with only one crossing. The only message it receives is "it's a wall" if it runs into walls, which does not incentivize it to explore more locations which will eventually lead it to the goal. In contrast, the position bonus easily solves these tasks. %Interestingly, the message bonus still succeeds on some other tasks which appear navigation based, such as \texttt{MultiRoom-N6-Lava} and \texttt{MultiRoom-N6-LavaMonsters}. On these tasks, the agent still receives one of two messages when it opens a door ("The door resists!" or "The door opens."), which can provide it with positive reward and encourage to open all the doors, which in turn guides it to the goal. We also created a variant of the \texttt{MultiRoom-N6-Lava} called \texttt{MultiRoom-N6-Lava-OpenDoor} which is identical, except that all doors are open--this way, the agent does not get any messages when it crosses the doorways. On this task, NovelD with the message bonus fails on all seeds. This illustrates how this variant can be sensitive to details of the task's construction. 

The symbolic image bonus performs poorly on the skill-based tasks, but performs well on some of the navigation-based ones. This is likely because for many of these tasks, the agent it the only moving entity, and hence the positional bonus and the image bonus will be similar. However, this is not always true. For example, in the \texttt{Labyrinth-Big} task, the environment is initially mostly hidden and gets revealed over time as the agent explores. This means that there is not a one-to-one correspondence between symbolic images and agent positions, as would be the case if the entire map was initially revealed. On this task, the position bonus succeeds but the symbolic bonus does not. Another particularly revealing example is the fact that the symbolic bonus succeeds on the \texttt{MultiRoom-N6-Lava-OpenDoor} environment, but fails on the \texttt{MultiRoom-N6-Lava} environment. The only difference between these two tasks is that in the former, all doors are initially open, which is marked by the \texttt{-} symbol. In the latter, they are initially closed (marked by the \texttt{+} symbol), and as they are opened by the agent, the symbol switches to \texttt{-}. In the open door version, there is a one-to-one correspondence between agent positions and symbolic images. On the closed door version, the number of possible symbols is the number of positions times the number of possible combinations of doors being open and closed---a much larger number. Once any door is opened, if the agent revisits a position it visited when the door was closed, it will once again receive bonus. This encourages the agent to revisit previously visited locations each time a door is opened, which does not align well with the task. This explains the poor performance of the symbolic image bonus on the closed door version of the task. This again illustrates how count-based episodic bonuses can be sensitive to slight changes in the task's construction.

% A particularly interesting insight into its behavior can be seen by comparing the results on \texttt{MultiRoom-N6-Lava} and \texttt{MultiRoom-N6-LavaMonsters} and \texttt{MultiRoom-N6-Lava-OpenDoor}. On \texttt{MultiRoom-N6-Lava-OpenDoor}, it performs almost identically to the \texttt{counts-position} bonus. Indeed, on this task the only changing entity is the agent and thus there is a close correspondence between different symbol images and agent positions. In \texttt{MultiRoom-N6-Lava}, the door symbol changes from \texttt{"+"} to \texttt{"-"} when it is opened, which increases the number of possible symbol images for each agent position by a factor equal to the number of doors, which is $5$. We see that in this environment, NovelD with the symbolic image bonus learns considerably slower than with \texttt{counts-position}. Finally, in \texttt{MultiRoom-N6-LavaMonsters} there are multiple monsters which move around the environment, creating a combinatorially large number of possible symbol images. On this task, NovelD with the symbolic image bonus fails on all seeds, while NovelD with the position bonus succeeds. This again illustrates how count-based episodic bonuses can be sensitive to slight changes in the task's construction. 

When using the count-based bonus based on the full observation, which does not make any assumptions about which parts of it are useful for the task, NovelD fails on all the tasks except for two. A close look at the stats vector (see Figure \ref{fig:modalities}) reveals that it contains a time counter which increments each time step: this effectively makes each observation unique, and hence the episodic bonus is constant \footnote{This is a simplification: there are some exceptions where the time counter is not incremented, such as if the agent runs into a wall or is searching through its inventory. This counts as a time step for the RL environment wrapper, but not for the underlying NetHack game engine. However these instances are relatively rare.}.

\subsection{Sensitivity of $\lambda$ regularizer}
\label{appendix:ridge_comparison}

\begin{figure}[h]
    \centering
    \includegraphics[width=\textwidth]{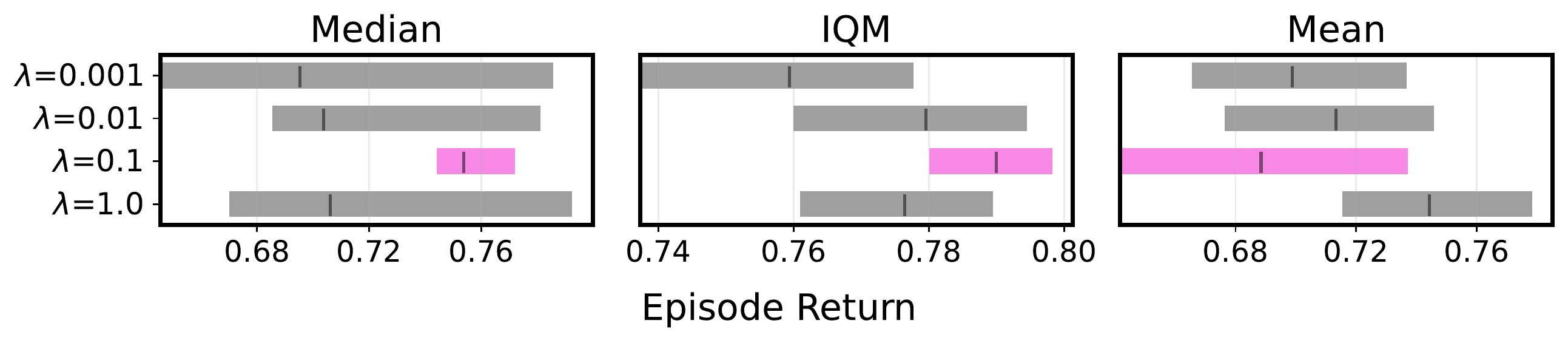}
    \caption{MiniHack performance for different values of $\lambda$ parameter. All MiniHack experiments in the paper use the value $\lambda=0.1$ unless otherwise noted. Intervals are computed using $5$ random seeds using stratified bootstrapping.}
    \label{fig:ridge_comparison}
\end{figure}

Figure \ref{fig:ridge_comparison} shows \alg's performance on MiniHack for different values of the covariance regularizer $\lambda$. There is no statistically significant difference between $\lambda=0.1$, which is the value we used in our experiments, and $\lambda=0.01$ or $\lambda=1.0$. For $\lambda=0.001$, we observe a statistically significant drop in performance for the IQM metric only. This shows that \alg~is fairly robust to the value of $\lambda$. 

\end{document}